\documentclass[runningheads]{llncs}

\usepackage{eccv}

\usepackage{eccvabbrv}

\usepackage{graphicx}
\usepackage{booktabs}
\usepackage[table]{xcolor}
\usepackage{multirow}
\usepackage{makecell}

\usepackage[accsupp]{axessibility}  

\usepackage[pagebackref,breaklinks,colorlinks,citecolor=eccvblue]{hyperref}
\usepackage{hyperref}
\newcommand{\mainref}[1]{#1}

\usepackage{orcidlink}

\begin{document}

\title{Tessellating The Earth} 

\titlerunning{Tessellating The Earth}

\author{
Daniel Cher \and
Hamza Iqbal \and
Eric Xing \and
Brian Wei \and
Nathan Jacobs 
}

\authorrunning{D.~Cher et al.}

\institute{Washington University in St. Louis \\
\email{\{cher, h.iqbal, e.xing, b.j.wei, jacobsn\}@wustl.edu}}

\maketitle

\begin{abstract}
Geolocation encoders, which map geographic coordinates to learned representations, are emerging as an effective means of capturing visual and non-visual characteristics from a latitude-longitude pair alone. However, existing approaches project coordinates onto fixed bases (e.g., spherical harmonics), allocating representational capacity uniformly and devoting equal resources to the open ocean and to a developing city. We introduce Tessellating the Earth (TTE), a location encoder built from learnable Spherical Voronoi partitions that concentrates representational capacity where it is needed in a fully differentiable, end-to-end manner. Each Voronoi site carries its own embedding and migrates during training toward discriminative areas. To bridge the gap between local spatial structure and global semantic understanding, we introduce \emph{global semantic tokens}: a set of shared learnable concept tokens that distill semantic knowledge from the satellite imagery into a compact vocabulary the location encoder can reference at inference, enabling geographically distant sites covering similar environments to share semantics. TTE sets a new state of the art for location encoders across a suite of geospatial classification and regression tasks, and achieves the strongest results when used as a geographic prior for fine-grained species classification on iNaturalist-2018. Code, and weights are available at \url{https://github.com/mvrl/TTE}.
\end{abstract}
\section{Introduction}
\label{sec:intro}

\begin{figure*}[t]
    \centering
    \begin{minipage}[t]{0.33\linewidth}
        \centering
        \includegraphics[width=\linewidth]{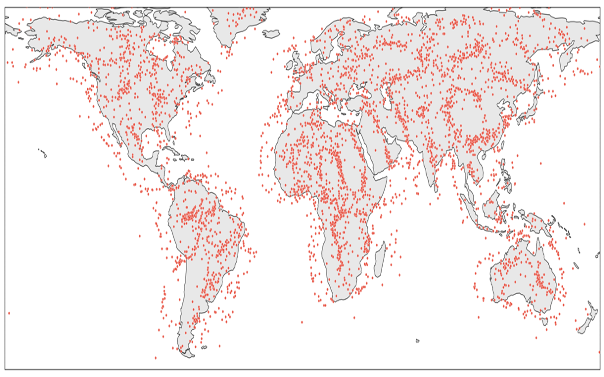}
        {\scriptsize (a) Learned site positions}
    \end{minipage}%
    \hfill
    \begin{minipage}[t]{0.33\linewidth}
        \centering
        \includegraphics[width=\linewidth]{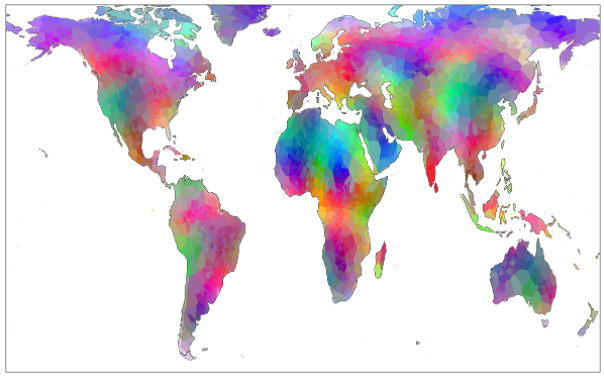}
        {\scriptsize (b) Site embeddings (ICA)}
    \end{minipage}%
    \hfill
    \begin{minipage}[t]{0.33\linewidth}
        \centering
        \includegraphics[width=\linewidth]{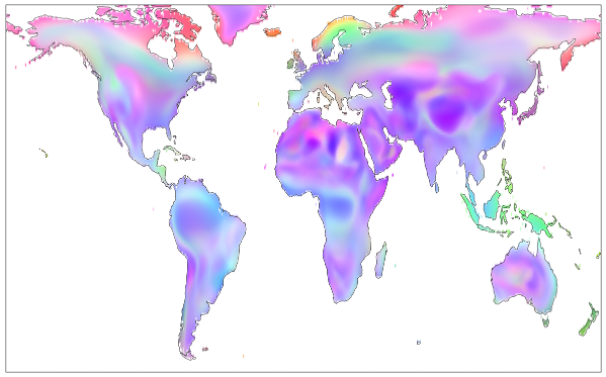}
        {\scriptsize (c) Location embeddings (ICA)}
    \end{minipage}
    \caption{\textbf{Learned spatial decomposition.} TTE partitions the sphere into learnable Voronoi regions that adapt to geographic complexity. (a)~After training, sites concentrate on land masses, clustering densely along coastlines and ecological boundaries while leaving oceans sparsely covered. (b)~Each site carries a learned embedding, with neighboring sites sharing similar embeddings (as indicated by color proximity), thereby forming a coherent spatial code. (c)~The final location embeddings produced by TTE vary smoothly across space, reflecting underlying geographic and ecological structure.}
    \label{fig:tessellation}
\end{figure*}

Where something is on Earth says a great deal about what is there. Climate, ecology, land cover, and built infrastructure are all closely tied to location, and all are highly predictive for a wide range of geospatial tasks. Location encoders aim to capture these latent properties by learning mappings from latitude–longitude coordinates to high-dimensional representations~\cite{klemmer2025earth}. These representations benefit applications including species distribution modeling~\cite{cole2023sinr,mac2019presence}, crop yield estimation~\cite{tseng2025galileo}, air quality forecasting~\cite{karimzadeh2025pm25}, canopy height estimation~\cite{lang2023canopy}, and satellite image synthesis~\cite{sastry2024geosynth,cher2026vectorsynth}. Once trained, they provide a compact representation of any location from coordinates alone, without requiring imagery at inference.

Recent work in building location encoders has focused on contrastive alignment between location and images. SatCLIP~\cite{klemmer2025satclip} aligns coordinates with satellite images, while GeoCLIP~\cite{cepeda2023geoclip} uses ground-level photographs. In each case, a positional encoding projects the raw coordinates into a high-dimensional feature space, and a neural network maps these features into an embedding that is trained to match the co-located image embedding. The choice of positional encoding is critical. Without explicit high-frequency structure, neural networks default to smooth, low-frequency solutions~\cite{rahaman2019spectral}, making raw coordinates insufficient for fine-grained spatial variation. Existing methods address this with fixed bases, whether spherical harmonics~\cite{klemmer2025satclip,russwurm2024siren}, multi-scale Fourier features~\cite{mai2020space2vec,mai2023sphere2vec}, or random Fourier features~\cite{cepeda2023geoclip}. These bases are mathematically convenient but share a fundamental limitation: their spatial structure is determined before training and cannot adapt to the data. Spherical harmonics, for instance, distribute capacity uniformly across the sphere, devoting as many basis functions to the open ocean as to a metropolitan city. As we show in \cref{sec:ablation}, replacing a learned spatial decomposition with a fixed one significantly reduces performance, confirming that this limitation is not merely theoretical.

\begin{figure}
    \centering
    \includegraphics[width=0.85\linewidth]{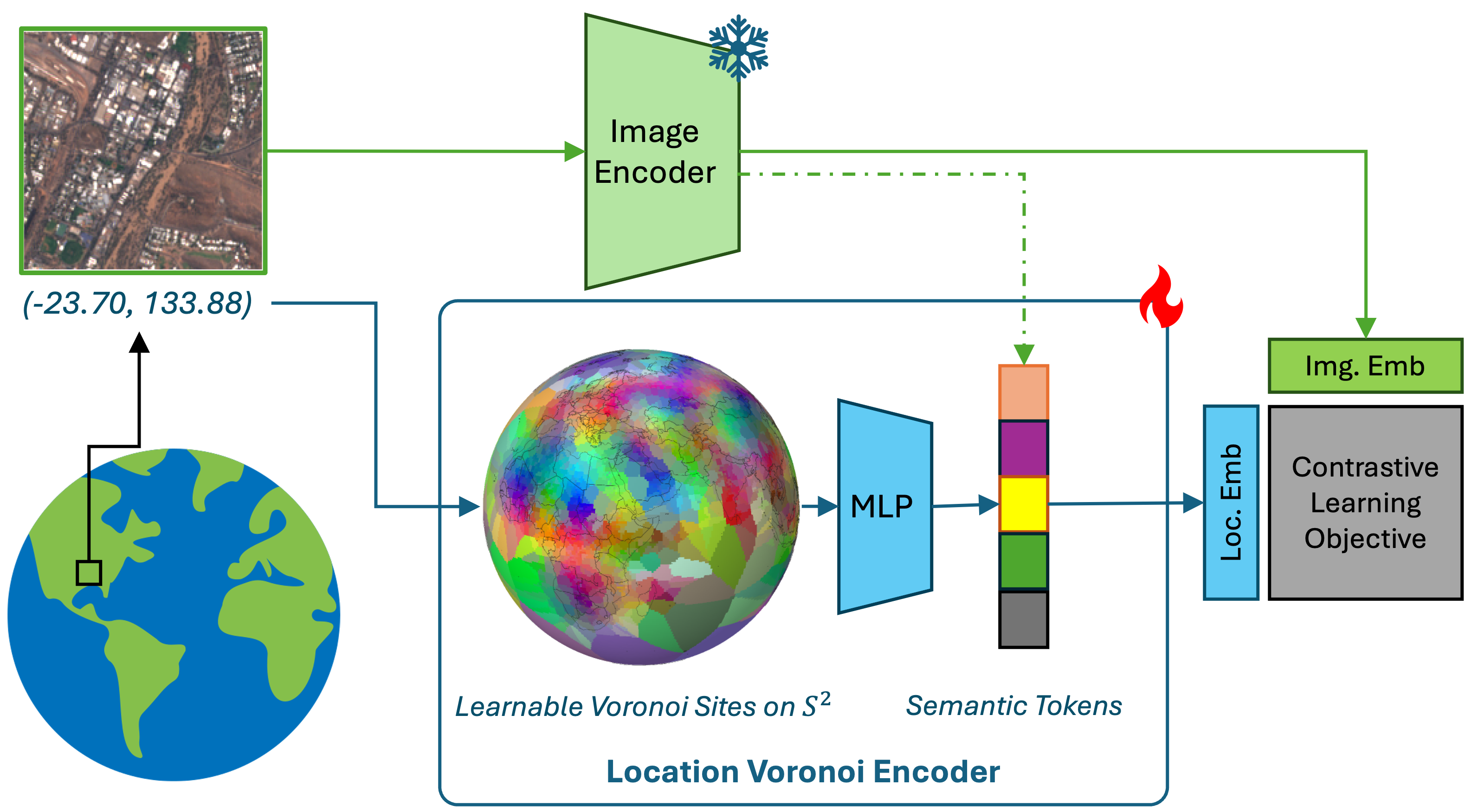}
    \caption{\textbf{Overview of TTE.} The \emph{location pathway} (bottom) maps a coordinate onto~$\mathbb{S}^{2}$, where learnable Voronoi sites perform soft assignment; the resulting embedding attends over shared \emph{semantic tokens} to produce the location embedding. The \emph{image pathway} (top) encodes the co-located satellite image with a frozen ViT. The resulting image embedding enters the contrastive objective directly and supervises the tokens during training (dashed). At inference only the location pathway is required.}
    \label{fig:method}
\end{figure}

We propose to replace these fixed bases with a learnable spatial decomposition. A Voronoi tessellation divides a surface into regions based on proximity to a set of generating sites, a construction used widely across computational geometry~\cite{okabe2000spatial, aurenhammer1991voronoi}, climate modeling~\cite{ringler2008multiresolution,skamarock2012mpas}, and neural scene representation~\cite{rebain2021derf}. \emph{Tessellating the Earth} (TTE) encodes locations through Spherical Voronoi partitions~\cite{disario2025sv} on the unit sphere using a set of learnable sites, each carrying its own embedding, whose soft assignment weights define how nearby coordinates are encoded (see \cref{fig:tessellation} for visualization of the learned model). Both site positions and embeddings are learnable parameters, jointly optimized through contrastive pretraining with satellite imagery. The contrastive objective drives sites to arrange themselves such that the resulting location embeddings discriminate between visually distinct places (see \cref{fig:method} for an architecture overview).

Spatial partitioning alone does not address semantic understanding. Our Spherical Voronoi representation assigns each site a local embedding, but two sites covering tropical forest on different continents have no explicit pathway to share or align their representations. To address this, we observe that the pretrained image encoder contains rich semantic knowledge about land types, but this information is only available during training. We introduce \emph{global semantic tokens}: a set of shared learnable concept tokens that distill this visual semantic knowledge into a compact vocabulary the location encoder can reference at inference time, when no imagery is available (see token spatial coherence \cref{fig:token_qualitative_quad}). This factors the problem into local spatial structure (site embeddings) and global semantic alignment (tokens), letting distant sites covering similar environments share capacity through common concepts.

Our contributions are:
\begin{itemize}
\item \textbf{Learnable Spherical Voronoi partitions for geographic encoding.} We introduce a location encoder whose spatial decomposition is learned end-to-end from visual supervision, replacing fixed bases with data-driven site placement. TTE sets a new state of the art among parametric location encoders across a suite of geospatial classification and regression benchmarks.
\item \textbf{Global semantic tokens for cross-modal alignment.} We propose shared learnable concept tokens that distill semantic knowledge from the pretrained image encoder, bridging the representational gap between coordinates and imagery. We show that they learn coherent visual semantic themes, drive quantitative improvements, and achieve the strongest results as a geographic prior for fine-grained species classification on iNaturalist-2018.
\end{itemize}
\section{Related Work}

\paragraph{Location encoders and geo-visual pretraining.}
Geographic coordinate encoding has progressed from geometry-aware transforms (e.g., wrap encodings)~\cite{mac2019presence} and soft grid-based methods~\cite{yin2019gps2vec} through multi-scale positional encodings in $\mathbb{R}^n$~\cite{mai2020space2vec,mai2023sphere2vec} to bases defined directly on the sphere~\cite{mai2022review}. Ru\ss{}wurm et al.\ couple spherical harmonics, an orthogonal basis on $S^2$, with sinusoidal representation networks (SIREN)~\cite{sitzmann2020siren}, providing tunable resolution via the maximum harmonic degree~\cite{russwurm2024siren}. Many of these encoders have been trained via contrastive alignment with geotagged imagery. SatCLIP applies CLIP-style pretraining on globally sampled Sentinel-2 imagery~\cite{klemmer2025satclip}, GeoCLIP frames worldwide geo-localization as image--location retrieval using hierarchical random Fourier feature encodings~\cite{cepeda2023geoclip}, and CSP learns location representations by aligning a location encoder with a frozen image encoder on ground-level and overhead imagery~\cite{mai2023csp}. These methods vary in imagery source and positional basis, but none adapt their spatial support to the underlying signal. A separate line of work addresses the limitations of contrastive alignment itself rather than the positional encoding. RANGE augments the location embeddings with image features retrieved from an external database at inference time~\cite{dhakal2025range}. While effective, this couples the encoder's performance to database coverage and introduces deployment constraints. Our goal is instead to improve the parametric location encoder itself, without external retrieval at test time.

\paragraph{Sphere-native representations and learnable partitions.}
Neural networks exhibit a well-documented spectral bias toward low frequencies~\cite{rahaman2019spectral}, motivating Fourier feature mappings~\cite{tancik2020fourier} and periodic activations~\cite{sitzmann2020siren} to lift coordinates into high-dimensional spaces before processing. On the sphere, this role is filled by spherical harmonics~\cite{russwurm2024siren}, multi-scale sinusoidal encodings~\cite{mai2020space2vec,mai2023sphere2vec}, and random Fourier features~\cite{cepeda2023geoclip}. Despite strong performance, these encoders fix their spatial support \emph{a priori}, which introduces several limitations. Spherical harmonics distribute capacity uniformly across the sphere, including over oceans ($\sim$71\% of the surface), and the number of coefficients grows quadratically with the maximum harmonic degree, making high-resolution representations expensive while still degrading near sharp spatial transitions~\cite{gelb1997gibbs,russwurm2024siren}. Cai and Balestriero~\cite{cai2025nolocation} confirm that this uniformity leads to systematic performance disparities, with coastlines and small landmasses consistently underserved. Localized alternatives such as spherical Gaussians~\cite{wang2009allfrequency} offer compact support but can be sensitive to initialization and yield weak gradients when primitives are misaligned~\cite{disario2025sv}. Concurrent work by Rao et al.~\cite{rao2026slepian} pursues localization differently, using Slepian functions as an analytically concentrated basis that achieves high resolution within a predefined region of interest. Both directions improve on uniform bases by focusing capacity locally, but each fixes that focus in advance: spherical Gaussians through their initial placement and Slepian functions through a region that must be specified before training, which limits the latter to localized prediction tasks. Di~Sario et al.\ instead introduce differentiable Spherical Voronoi (SV) partitions with learnable sites and per-site temperatures, where softmax assignment provides non-zero gradients to every site and enables fully data-adaptive capacity allocation~\cite{disario2025sv}. DeRF~\cite{rebain2021derf} earlier demonstrated the principle of learnable Voronoi decomposition for radiance fields, showing that learned partitions outperform fixed grids for heterogeneous scenes. TTE adapts SV from directional appearance modeling to geographic representation learning and extends it with global semantic tokens for cross-modal pretraining.

\paragraph{Global semantic tokens.}
Several architectures use small sets of learnable tokens, not tied to any input, as shared communication channels for aggregation across local components. Many popular generalist vision architectures, from the original ViT~\cite{dosovitskiy2020image} to modern VLM architectures~\cite{Qwen3-VL, alayrac2022flamingo, jaegle2021perceiver, li2023blip2, dai2023instructblip} use attention mechanisms combined with learnable queries to distill sequence information. Darcet et al.\ show that adding explicit register tokens to Vision Transformers provides dedicated global memory and eliminates feature-map artifacts caused by repurposing input patches for implicit storage~\cite{darcet2024registers}, while Set Transformer~\cite{lee2019settransformer} uses inducing points that serve a similar role. MetaEmbed~\cite{xiao2025metaembed} extends learnable queries as a flexible way to scale late interaction for retrieval. TTE's global semantic tokens share the learnable-token mechanism but serve a distinct purpose. Rather than cross-attending within a single encoder, they act as shared latent anchors between two modalities, distilling semantic knowledge from the pretrained image encoder into a compact vocabulary that the location encoder can reference at inference time, enabling information exchange across distant Voronoi cells while preserving the locality benefits of a partition-based encoding.
\section{Methodology}
\label{sec:method}

\begin{figure*}
    \centering
    \small

    \begin{minipage}[t]{0.48\linewidth}
        \centering
        \includegraphics[width=0.32\linewidth]{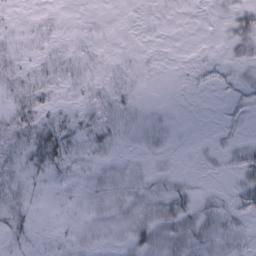}%
        \hfill
        \includegraphics[width=0.32\linewidth]{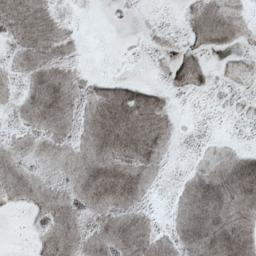}%
        \hfill
        \includegraphics[width=0.32\linewidth]{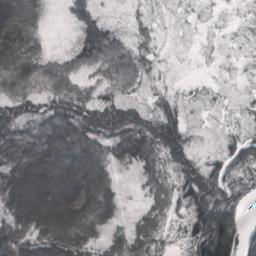}
        \vspace{2pt}
        \includegraphics[width=\linewidth]{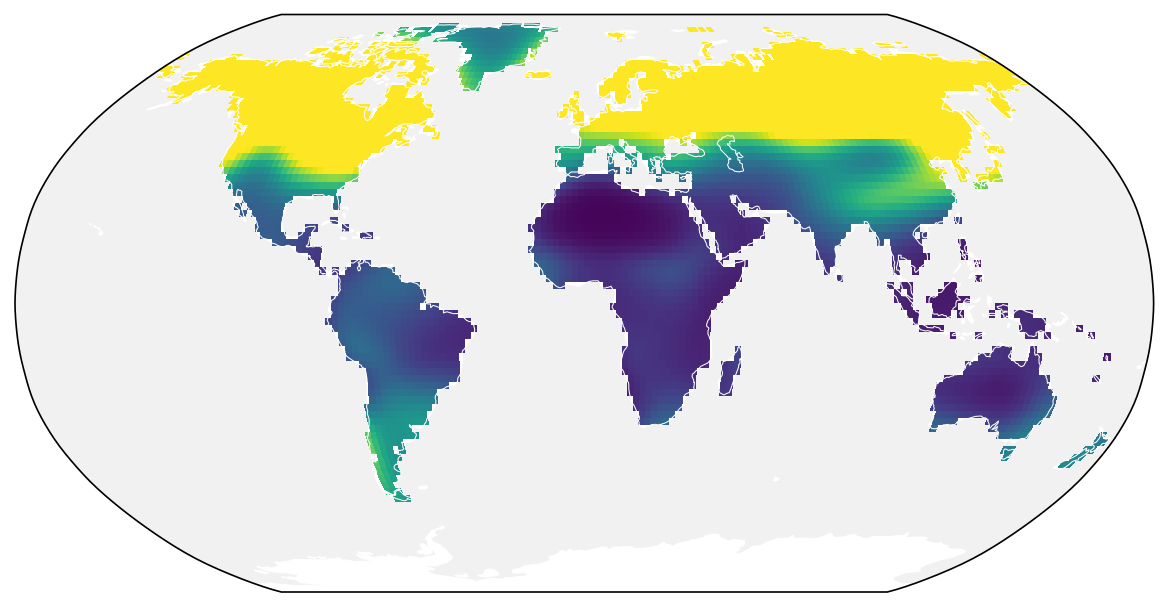}
        \vspace{1pt}
        {\scriptsize (a) Snow \& ice cover}
    \end{minipage}
    \hfill
    \begin{minipage}[t]{0.48\linewidth}
        \centering
        \includegraphics[width=0.32\linewidth]{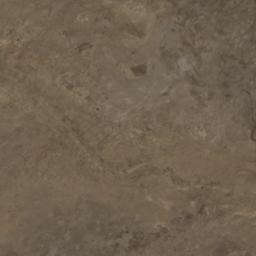}%
        \hfill
        \includegraphics[width=0.32\linewidth]{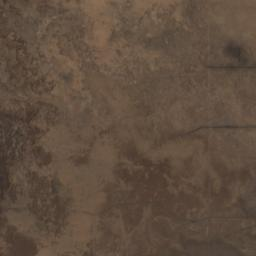}%
        \hfill
        \includegraphics[width=0.32\linewidth]{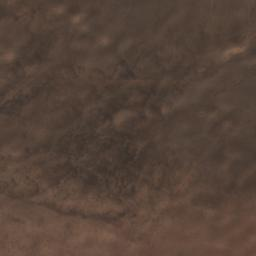}
        \vspace{2pt}
        \includegraphics[width=\linewidth]{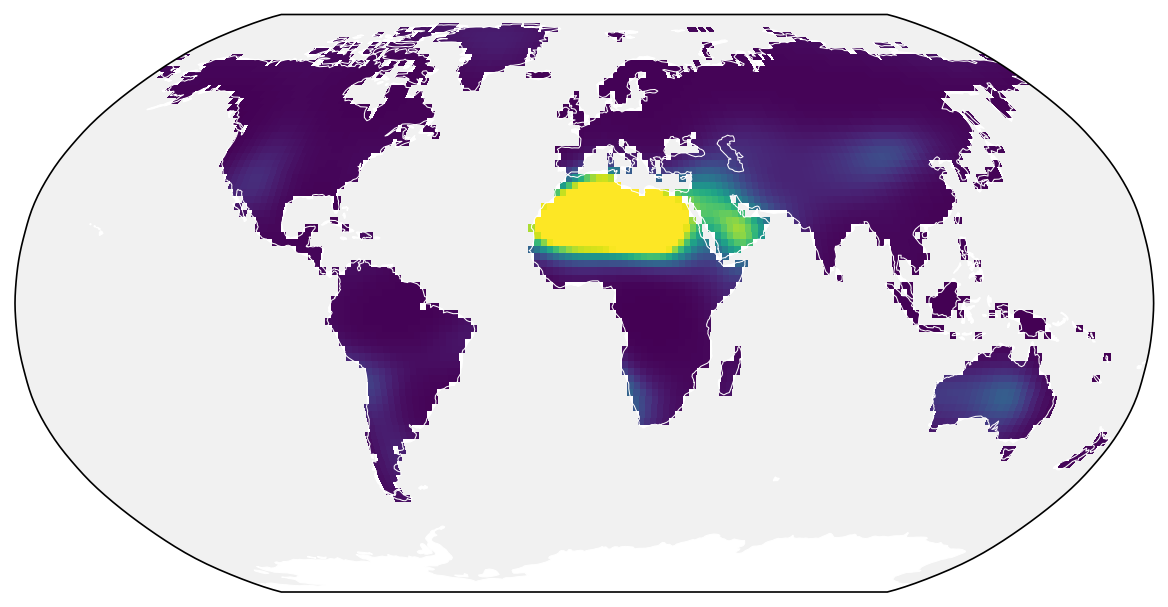}
        \vspace{1pt}
        {\scriptsize (b) Arid / desert}
    \end{minipage}

    \vspace{10pt}

    \begin{minipage}[t]{0.48\linewidth}
        \centering
        \includegraphics[width=0.32\linewidth]{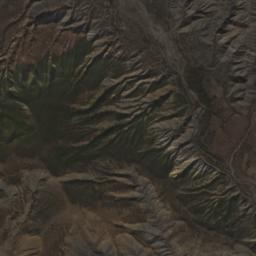}%
        \hfill
        \includegraphics[width=0.32\linewidth]{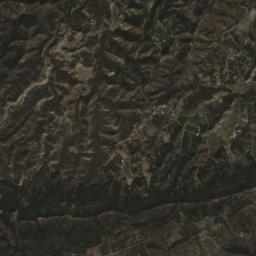}%
        \hfill
        \includegraphics[width=0.32\linewidth]{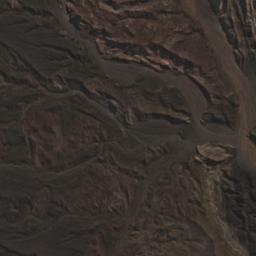}
        \vspace{2pt}
        \includegraphics[width=\linewidth]{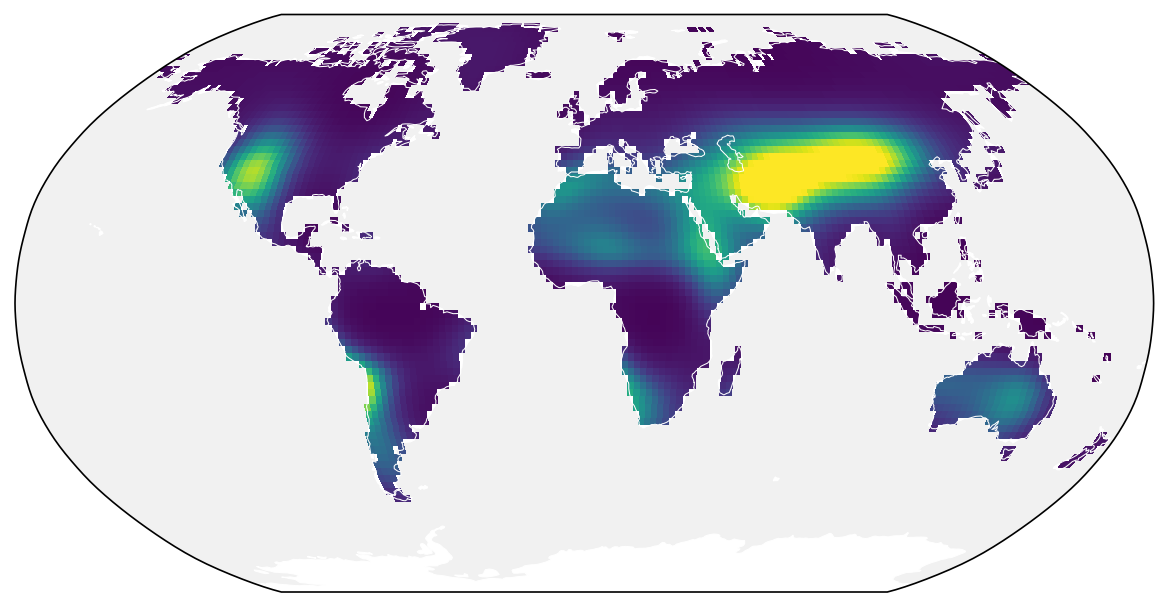}
        \vspace{1pt}
        {\scriptsize (c) Rocky terrain}
    \end{minipage}
    \hfill
    \begin{minipage}[t]{0.48\linewidth}
        \centering
        \includegraphics[width=0.32\linewidth]{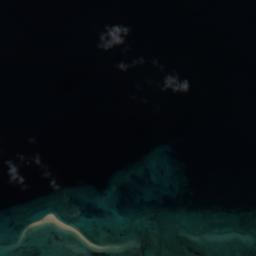}%
        \hfill
        \includegraphics[width=0.32\linewidth]{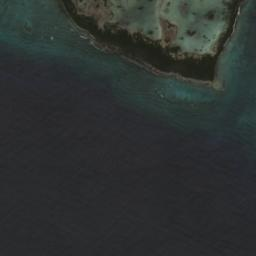}%
        \hfill
        \includegraphics[width=0.32\linewidth]{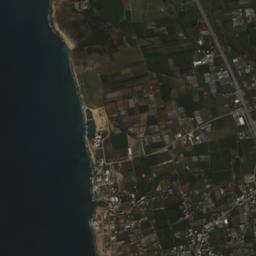}
        \vspace{2pt}
        \includegraphics[width=\linewidth]{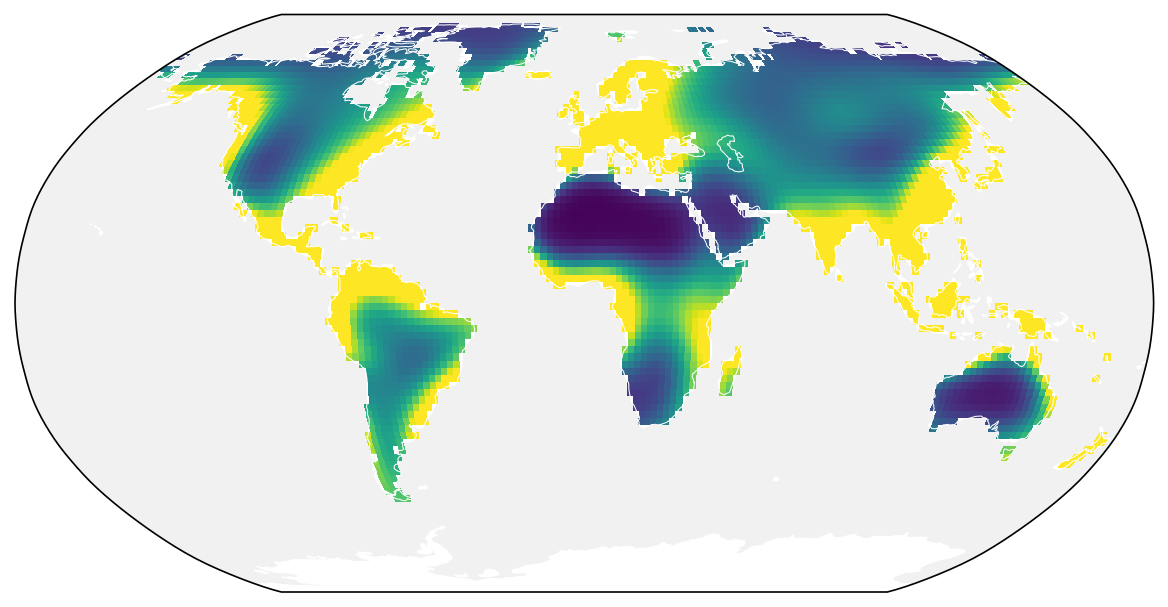}
        \vspace{1pt}
        {\scriptsize (d) Coastal regions}
    \end{minipage}

    \caption{\textbf{Learned global semantic tokens.} Each sub-figure displays a specific semantic token with three top-attending satellite image patches (top) and the corresponding global spatial attention map (bottom). The visualization demonstrates that the model's internal tokens naturally specialize in distinct geographic and climatic features without explicit supervision.}
    \label{fig:token_qualitative_quad}
\end{figure*}

An overview of TTE is shown in \cref{fig:method}, with additional details on the global semantic tokens provided in \cref{fig:token_mechanism}. The model has three components: a Spherical Voronoi encoder that maps coordinates to embeddings (\cref{sec:prelim}), semantic tokens that enable global semantic sharing across localized sites (\cref{sec:registers}), and a set of training objectives that jointly optimize spatial and semantic representations (\cref{sec:training}).

\subsection{Preliminaries: Spherical Voronoi Partitions}
\label{sec:prelim}

We build on the differentiable Spherical Voronoi representation introduced by Di Sario~\etal~\cite{disario2025sv} for view-dependent appearance modeling. Given $K$ sites $\{\mathbf{s}_1, \ldots, \mathbf{s}_K\}$ on the unit sphere $S^2$, the soft assignment weight of a query point $\mathbf{x} \in S^2$ to site $k$ is:
\begin{equation}
\label{eq:voronoi_weight}
w_k(\mathbf{x}) = \frac{\exp(\tau_k \cdot (\mathbf{s}_k \cdot \mathbf{x}))}{\sum_{j=1}^{K} \exp(\tau_j \cdot (\mathbf{s}_j \cdot \mathbf{x}))}
\end{equation}
where $\tau_k > 0$ is a per-site learnable temperature controlling partition sharpness. Since both $\mathbf{s}_k$ and $\mathbf{x}$ are unit vectors on $S^2$, their dot product equals the cosine of the angular distance between them, providing a natural proximity measure on the sphere. The softmax formulation ensures well-defined gradients for all sites and avoids competition between overlapping kernels that other spherical representations exhibit~\cite{disario2025sv}.

We adapt this formulation for geographic encoding by replacing the scalar site values of~\cite{disario2025sv} with learnable embedding vectors $\mathbf{e}_k \in \mathbb{R}^D$. The location embedding for a query $\mathbf{x}$ is the soft combination:
\begin{equation}
\label{eq:location_embedding}
\mathbf{f}(\mathbf{x}) = \sum_{k=1}^{K} w_k(\mathbf{x}) \cdot \mathbf{e}_k
\end{equation}

We use $K = 4{,}096$ sites, each producing a 384-dimensional embedding, which is then processed by a two-block residual MLP (linear projection to 512 dimensions, two residual blocks with ReLU and 0.5 dropout) to produce the final 512-dimensional location embedding $\mathbf{f}_{\text{out}}$. Sites are initialized on a Fibonacci lattice filtered to land areas (excluding Antarctica), with temperatures initialized to $\tau_k = 45$. All parameters, including site positions $\mathbf{s}_k$, temperatures $\tau_k$, and embeddings $\mathbf{e}_k$, are jointly optimized~\cite{loshchilov2017decoupled_adamw} end-to-end. Positions are re-normalized to $S^2$ after each gradient step.

\subsection{Global Semantic Tokens}
\label{sec:registers}

\begin{figure}
    \centering
    \includegraphics[width=.9\linewidth]{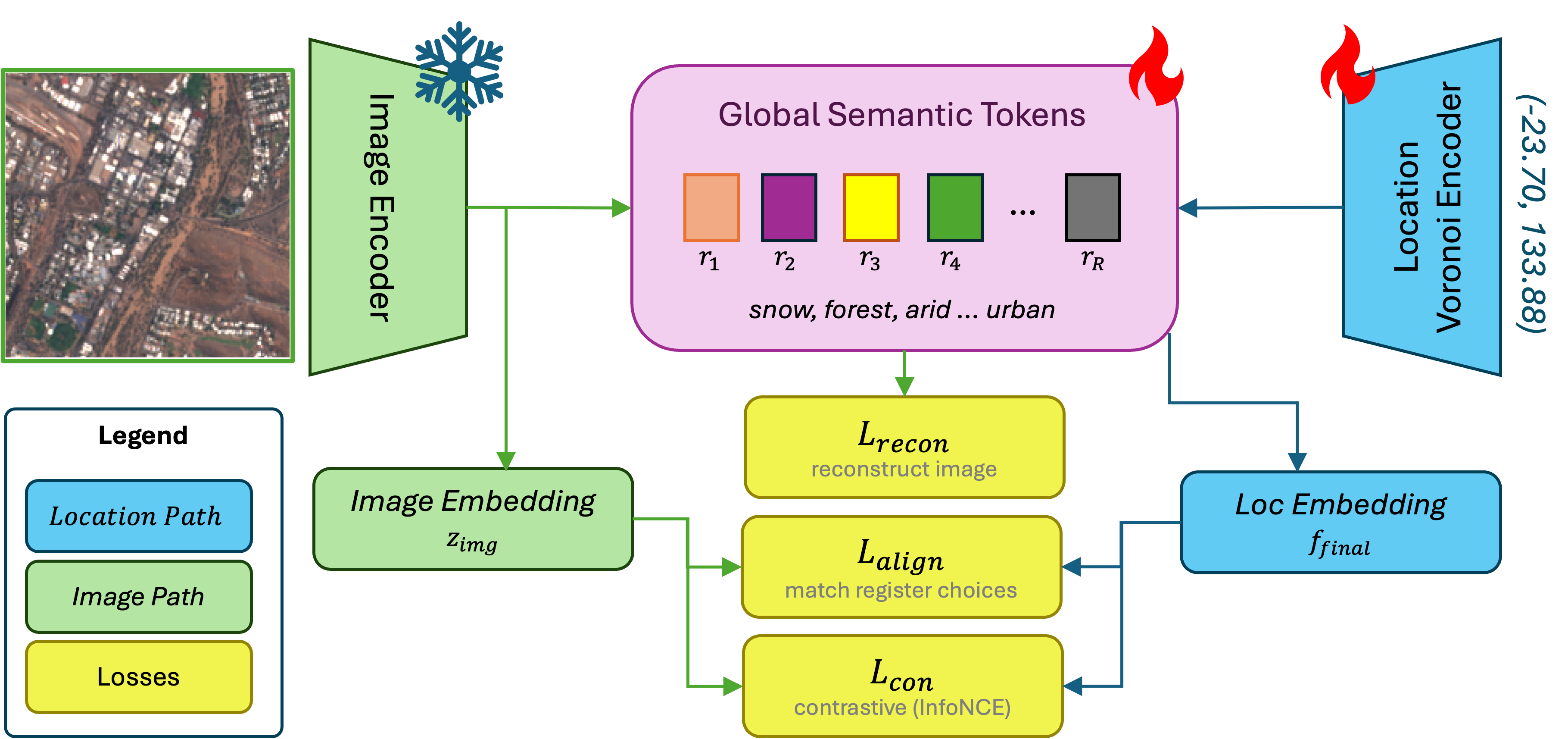}
    \caption{\textbf{Global semantic token mechanism.} Both paths attend to shared learnable tokens. The image path (training only) produces peaked attention via a fixed low temperature; the location path learns to match this distribution. Three losses jointly optimize the system.}
    \label{fig:token_mechanism}
\end{figure}

A visualization of the token alignment is shown in \cref{fig:token_mechanism}. Each Voronoi site carries an embedding that captures the representation of its local region, but sites are spatially isolated. A site covering tropical forest in South America has no mechanism to share meaning with a site covering tropical forest in Central Africa. The pretrained image encoder contains rich semantic knowledge about land types, but this information is only available during training. Global semantic tokens address both limitations by distilling visual semantics into a compact set of learnable concept tokens that the location encoder can reference at inference time, providing a global layer of shared concepts that all sites can access.

We define $R$ learnable concept tokens $\{\mathbf{r}_1, \ldots, \mathbf{r}_R\} \subset \mathbb{R}^{512}$, initialized as orthogonal unit vectors. Both the location encoder and image encoder compute attention distributions over these tokens.

\paragraph{Token attention.}
Both the location encoder and image encoder attend to the tokens through the same mechanism: a linear projection to $R$-dimensional logits followed by temperature-scaled softmax. For the location path:
\begin{equation}
\label{eq:register_attn}
\mathbf{a}_{\text{loc}} = \text{softmax}\!\left((\mathbf{W}_{\text{loc}}\,\mathbf{f}(\mathbf{x}) + \mathbf{b}_{\text{loc}}) \,/\, T_{\text{loc}}\right), \quad
\mathbf{z}_{\text{loc}} = \sum_{r=1}^{R} a_{\text{loc},r} \cdot \mathbf{r}_r
\end{equation}
The image path is analogous, projecting image features $\mathbf{v}$ through a separate projection $\mathbf{W}_{\text{img}}$. The critical difference is temperature. The location path uses a learnable $T_{\text{loc}}$, while the image path uses a fixed low temperature $T_{\text{img}} = 0.05$ that produces peaked, nearly one-hot attention. Each image activates one or two tokens strongly, and the location path must learn to predict this peaked distribution from coordinates alone. This asymmetry provides a sharper alignment target than symmetric soft attention, where both paths tend toward diffuse distributions that weaken token specialization (\cref{tab:ablation}).

The final location embedding is an equal-weight combination of the token-attended representation and a linear projection of the Voronoi embedding:
\begin{equation}
\label{eq:blend}
\mathbf{f}_{\text{final}} = \text{LayerNorm}\!\left((1 - \alpha) \cdot \mathbf{W}\,\mathbf{f}(\mathbf{x}) + \alpha \cdot \mathbf{z}_{\text{loc}}\right)
\end{equation}
where $\mathbf{W}$ is a linear projection that maps the Voronoi embedding to the output dimensionality and $\alpha = 0.5$.

\subsection{Training Objective}
\label{sec:training}

The model is trained with losses that jointly optimize spatial encoding and semantic alignment.
\noindent
\paragraph{Contrastive loss.} The primary objective is a symmetric contrastive loss~\cite{radford2021learning, klemmer2025satclip} over batches of $N$ co-located image-location pairs. Let $\mathbf{V}, \mathbf{F} \in \mathbb{R}^{N \times 512}$ denote the L2-normalized image and location embeddings, and $\tau_{\text{logit}}$ a learnable temperature. The logit matrix is $\mathbf{L} = 
\tau_{\text{logit}} \cdot \mathbf{V} \mathbf{F}^\top$, and the loss is:
\begin{equation}
\mathcal{L}_{\text{con}} = \tfrac{1}{2} \left[ \text{CE}(\mathbf{L}, \mathbf{I}) + \text{CE}(\mathbf{L}^\top, \mathbf{I}) \right]
\end{equation}
where $\mathbf{I}$ is the identity matrix. 
\noindent
\paragraph{Reconstruction loss.}
The reconstruction loss requires that the tokens can reconstruct image features, preventing token collapse and ensuring the concept tokens form a useful basis covering the image feature space:
\begin{equation}
\label{eq:recon}
\mathcal{L}_{\text{recon}} = \frac{1}{N} \sum_{i=1}^{N} \left\|\mathbf{z}_{\text{img},i} - \mathbf{v}_i\right\|^2
\end{equation}
where $\mathbf{z}_{\text{img},i} = \sum_{r} a_{\text{img},r}^{(i)} \cdot \mathbf{r}_r$ is the token-attended image representation.

\paragraph{Alignment loss.}
The alignment loss encourages the location attention distribution over tokens to match the image attention distribution via KL divergence, with stop-gradient on the image side:
\begin{equation}
\label{eq:align}
\mathcal{L}_{\text{align}} = \frac{1}{N} \sum_{i=1}^{N} \text{KL}\!\left(\mathbf{a}_{\text{loc},i} \,\|\, \text{sg}(\mathbf{a}_{\text{img},i})\right)
\end{equation}
where $\text{sg}(\cdot)$ denotes stop-gradient. Gradients flow only to the location path with image attention serving as a fixed target derived from visual content. The location encoder learns to predict which semantic concepts are present at a coordinate from the coordinate alone.

The total loss objective is:
\begin{equation}
\label{eq:total_loss}
\mathcal{L} = \mathcal{L}_{\text{con}} + \lambda_{\text{recon}} \cdot \mathcal{L}_{\text{recon}} + \lambda_{\text{align}} \cdot \mathcal{L}_{\text{align}}
\end{equation}
with $\lambda_{\text{recon}} = 100$ and $\lambda_{\text{align}} = 0.1$. 
The high reconstruction weight ensures strong semantic association, while the lower alignment weight allows location attention to deviate when geographically appropriate.
\section{Experiments}
\label{sec:exp}

We evaluate TTE on an extensive geospatial benchmark suite covering classification and regression tasks, as well as fine-grained species classification on iNaturalist-2018~\cite{inat2018}. Classification tasks include Biome and EcoRegion classification using the WWF ecoregion delineations~\cite{dinerstein2017ecoregion}, and Country classification. Regression tasks include gridded air temperature~\cite{hooker2018temperature}, elevation and population density~\cite{rolf2021mosaiks}, and California Housing prices derived from 1990 census data~\cite{pace1997housing}. Following prior work, we train a linear probe on top of inferenced coordinate embeddings for each downstream task and model. For contrastive pretraining, we follow the SatCLIP~\cite{klemmer2025satclip} protocol using globally sampled Sentinel-2 imagery, and use the same MoCo-pretrained image encoder~\cite{wang2022ssl4eo} across all methods for fair comparison. In addition, we analyze inference-time augmentation strategies in the appendix.

\begin{table*}
\centering
\small
\caption{\textbf{Geospatial Benchmarks.} Comparison across geospatial classification (accuracy) and regression ($R^2$) tasks. Classification tasks include Biome, EcoRegion and Country classification. Regression tasks include Temperature, Elevation, World Population and Cali-Housing estimation. Averages are computed within task type. Across $N{=}5$ training runs, TTE's results vary by at most $\pm0.3$ (classification accuracy) and $\pm0.009$ ($R^2$), well below the margins over competing methods.}
\setlength{\tabcolsep}{3pt}
\begin{tabular}{l|ccc|c|cccc|c}
\hline
& \multicolumn{3}{c|}{\textbf{Classification}} 
& 
& \multicolumn{4}{c|}{\textbf{Regression}} 
& \\
\cline{2-4} \cline{6-9}
\textbf{Method}
& Bio & Eco & Ctry & Cls $\uparrow$
& Temp & Elev & Pop & Cali & Reg $\uparrow$ \\
\hline
Direct & 29.1 & 0.6  & 66.9 & 32.2 & 0.381 & 0.025 & 0.053 & 0.238 & 0.174 \\
Cartesian 3D     & 30.2 & 1.8  & 66.9 & 33.0 & 0.362 & 0.030 & 0.162 & 0.240 & 0.199 \\
Wrap~\cite{mac2019presence} & 34.4 & 1.1  & 69.7 & 35.1 & 0.861 & 0.085 & 0.328 & 0.239 & 0.378 \\
Theory~\cite{gao2019learning} & 33.5 & 1.0  & 72.5 & 35.7 & 0.849 & 0.093 & 0.330 & 0.254 & 0.382 \\
SphereM~\cite{mai2023sphere2vec} & 36.4 & 27.3 & 72.7 & 45.5 & 0.629 & 0.139 & 0.302 & 0.423 & 0.373 \\
SphereM+~\cite{mai2023sphere2vec} & 58.7 & 50.1 & 76.1 & 61.6 & 0.886 & 0.294 & 0.421 & 0.543 & 0.536 \\
SphereC~\cite{mai2023sphere2vec} & 36.3 & 52.9 & 72.9 & 54.0 & 0.461 & 0.185 & 0.335 & 0.496 & 0.369 \\
SphereC+~\cite{mai2023sphere2vec} & 53.2 & 61.6 & 73.6 & 62.8 & 0.842 & 0.260 & 0.392 & 0.544 & 0.510 \\
CSP-INat~\cite{mai2023csp} & 61.1 & 57.1 & 75.9 & 64.7 & 0.717 & 0.388 & 0.554 & 0.462 & 0.530 \\
CSP-FMoW~\cite{mai2023csp} & 61.4 & 58.0 & 81.3 & 66.9 & 0.865 & 0.399 & 0.580 & 0.541 & 0.596 \\
SINR~\cite{cole2023sinr} & 67.9 & 54.9 & \textit{88.3} & 70.4 & \textit{0.942} & 0.644 & \textit{0.726} & 0.420 & 0.683 \\
GeoCLIP~\cite{cepeda2023geoclip} & 70.2 & \textit{71.6} & 81.3 & 74.4 & 0.916 & 0.604 & 0.698 & \textbf{0.708} & \textit{0.732} \\
SatCLIP~\cite{klemmer2025satclip} & 68.9 & 69.3 & 82.8 & 73.7 & 0.825 & \textit{0.666} & 0.684 & 0.400 & 0.644 \\
TaxaBind~\cite{sastry2025taxabind}         & \textit{72.3} & \textbf{72.9} & 86.4 & \textit{77.2} & 0.897 & 0.581 & 0.712 & \textit{0.684} & 0.719 \\
\hline
\rowcolor{gray!15}
TTE 
& \textbf{77.8} & 67.4 & \textbf{94.8} & \textbf{80.0}
& \textbf{0.946} & \textbf{0.839} & \textbf{0.790} & 0.532 & \textbf{0.777} \\
\hline
\end{tabular}
\label{tab:geospatial_eval}
\end{table*}

\subsection{Downstream Task Performance}

\begin{figure*}[t]
    \centering
    \includegraphics[width=.9\linewidth]{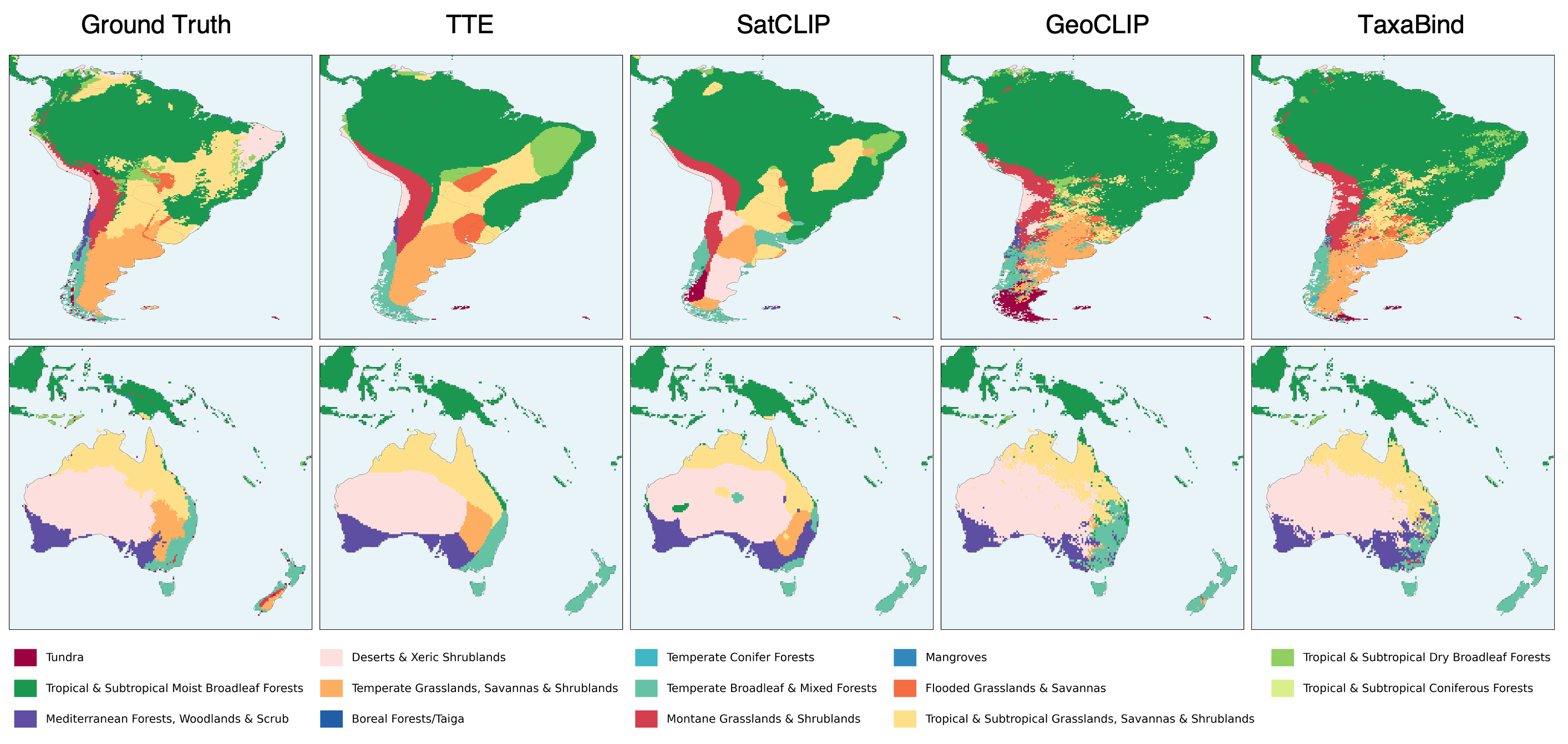}
    \caption{\textbf{Biome classification predictions.} Ground truth WWF biome labels (left) with predictions from a host of location encoders. We used a linear probe on each encoder's embeddings. TTE produces more spatially coherent predictions with smoother biome boundaries, while other models show more fragmentation within homogeneous regions.}
    \label{fig:biome_preds}
\end{figure*}

\cref{tab:geospatial_eval} presents results across the geospatial benchmark suite. TTE leads on a majority of individual tasks, with the largest margins on Biome (+5.5 over TaxaBind), Country (+6.5 over SINR), and Elevation (+0.173 over SatCLIP). \cref{fig:biome_preds} visualizes the spatial biome predictions from a linear probe, where TTE produces more spatially coherent regions compared to other approaches. The two tasks where TTE does not lead reflect known data alignment advantages of other methods. On EcoRegion, TaxaBind performs best, with its location encoder aligned to iNaturalist species observations and WorldClim bioclimatic variables that provide training signal directly matched to ecological boundaries. The California Housing uses 1990 census data, which creates a temporal mismatch between target labels and contemporary satellite imagery disproportionately negatively affecting encoders trained on Sentinel-2~\cite{dhakal2025range}, while methods trained on ground-level photographs with dense coverage of Western countries encode street-level cues more aligned with housing value. Despite these task-specific gaps, TTE achieves the highest average classification accuracy (80.0\%) and regression $R^2$ (0.777) among all parametric methods, improving over the previous best by 2.8\% and 0.045 respectively.

\begin{table}
\centering
\caption{\textbf{Geo-prior Benchmarking.} Top-$k$ classification accuracy ($\%$) on the iNaturalist-2018 test split. Incorporating geographic priors consistently improves fine-grained species classification over the image-only baseline. Our method (TTE) achieves the best performance across all $k$.}
\setlength{\tabcolsep}{4pt}
\begin{tabular}{l|cccc}
\hline
 & Top-1 & Top-3 & Top-5 & Top-10 \\
\hline
Img                & 66.1 & 83.3 & 88.0 & 92.2 \\
\hline
Img + CSP          & 72.9 & 87.9 & 91.6 & 94.8 \\
Img + GeoCLIP      & 72.9 & 88.2 & 91.9 & 95.2 \\
Img + CSP INat$^\ast$ & 74.4 & 88.8 & 92.2 & 94.9 \\
Img + SatCLIP      & 75.1 & 88.7 & 91.9 & 94.5 \\
Img + TaxaBind      & 75.1 & \textit{89.7} & \textit{93.0} & \textit{95.6} \\
\hline
\rowcolor{gray!15}
Img + TTE & \textbf{76.2} & \textbf{90.0} & \textbf{93.2} & \textbf{95.7} \\
\hline
\end{tabular}
\label{tab:inat_topk}
\end{table}

\cref{tab:inat_topk} evaluates whether self-supervised location representations generalize as a geographic prior for fine-grained species classification on iNaturalist-2018~\cite{inat2018}. Following~\cite{mac2019presence}, let $y$ denote the species label, $I$ the image, and $G$ the geographic location. We train a linear model on pre-trained geo-embeddings to produce a location-conditioned species distribution $P(y \mid G)$, and combine this with a pretrained image classifier's predictions $P(y \mid I)$ via $P(y \mid I, G) \propto P(y \mid I) \cdot P(y \mid G)$, under the conditional independence assumption of Mac~Aodha~\etal~\cite{mac2019presence}. TTE achieves the best Top-$k$ accuracy across all $k$, improving Top-1 from 66.1\% (image only) to 76.2\% and outperforming SatCLIP (75.1) and GeoCLIP (72.9). Notably, it also outperforms TaxaBind, whose location encoder is aligned with species data and trained on iNaturalist imagery. iNaturalist contains over 8{,}000 species classes whose distributions are governed by fine-grained environmental variation. With this many classes, even modest improvements in the geographic prior translate to meaningful gains in species-level predictions, as the prior effectively reweights the classifier's output across thousands of candidate species. The strong performance confirms that TTE's embeddings capture semantically rich geographic structure that transfers beyond the tasks seen during pretraining.

\begin{figure*}[b!]
    \centering
    \caption{\textbf{Ablation study.} \textit{Top:} performance vs.\ Voronoi site count (left) and token count $R$ (right). Performance peaks at 4{,}096 sites and $R = 64$ tokens. \textit{Bottom:} component ablations. Each row removes or modifies one component from the full model.}
    \begin{minipage}[t]{0.48\linewidth}
        \centering
        \includegraphics[width=\linewidth]{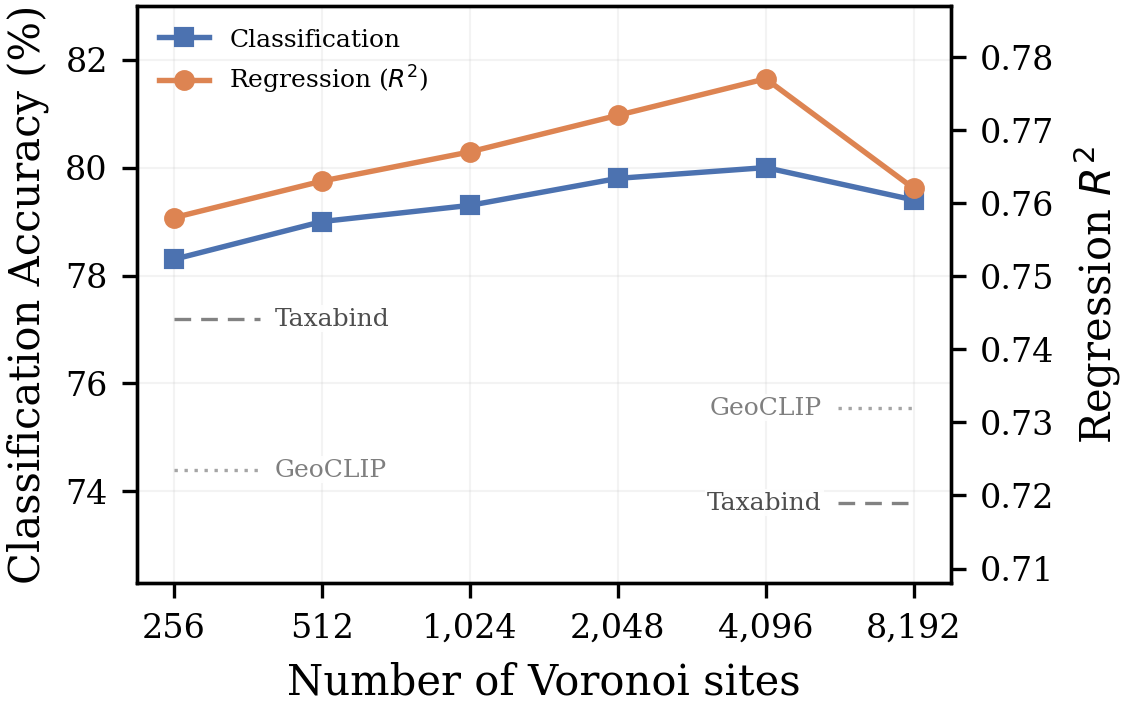}
    \end{minipage}%
    \hfill
    \begin{minipage}[t]{0.48\linewidth}
        \centering
        \includegraphics[width=\linewidth]{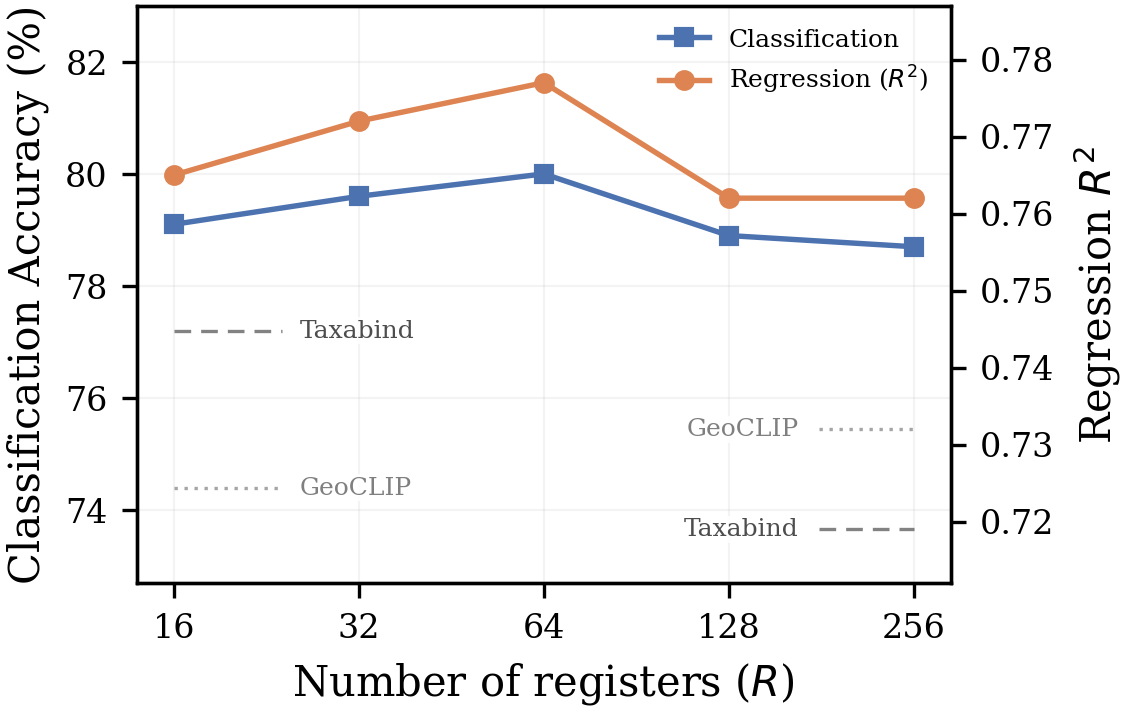}
    \end{minipage}

    \vspace{6pt}

    \scriptsize
    \setlength{\tabcolsep}{3pt}
    \begin{tabular}{l|ccc|c|cccc|c}
        \hline
        & Bio & Eco & Ctry & Cls $\uparrow$
        & Temp & Elev & Pop & Cali & Reg $\uparrow$ \\
        \hline
        \rowcolor{gray!15}
        \textbf{TTE (full)}
            & \textbf{77.8} & \textbf{67.4} & \textbf{94.8} & \textbf{80.0}
            & \textbf{0.946} & \textbf{0.839} & \textbf{0.790} & \textbf{0.532} & \textbf{0.777} \\
        \hline
        \multicolumn{10}{l}{\textit{Site ablations}} \\
        \quad fixed sites
            & 66.6 & 45.4 & 89.1 & 67.1
            & 0.936 & 0.644 & 0.727 & 0.299 & 0.652 \\
        \hline
        \multicolumn{10}{l}{\textit{Token ablations}} \\
        \quad w/o tokens
            & 74.2 & 62.2 & 92.6 & 76.3
            & 0.950 & 0.781 & 0.773 & 0.448 & 0.739 \\
        \quad w/o $\mathcal{L}_{\text{align}}$
            & 75.5 & 65.3 & 94.0 & 78.3
            & 0.950 & 0.822 & 0.784 & 0.496 & 0.763 \\
        \quad w/o $\mathcal{L}_{\text{recon}}$
            & 76.3 & 65.7 & 94.5 & 78.8
            & 0.952 & 0.828 & 0.784 & 0.499 & 0.766 \\
        \quad sym.\ soft att.\ ($T{=}0.2$)
            & 76.0 & 64.2 & 93.2 & 77.8
            & 0.952 & 0.794 & 0.776 & 0.484 & 0.752 \\
        \hline
    \end{tabular}
    \label{fig:ablation}
    \label{tab:ablation}
\end{figure*}

\subsection{Ablation Study}
\label{sec:ablation}

\paragraph{Component ablations.}
\cref{fig:ablation} isolates the contribution of each component. Fixing sites at their initial locations yields the largest decline among ablations ($-12.9\%$ classification, $-0.125$ regression), falling below SatCLIP on several tasks. This suggests that the ability of sites to migrate during training is critical, reinforcing that uniform spatial allocation is a practical bottleneck for location encoders, not merely a theoretical concern.

Among token ablations, removing the entire semantic branch (\textit{w/o tokens}) has the largest effect ($-3.7\%$ classification, $-0.038$ regression), with the steepest losses on Biome, EcoRegion, and Elevation. Removing individual losses produces smaller drops. \textit{w/o~$\mathcal{L}_{\text{align}}$} allows the location path to leverage semantic content but selects tokens less reliably without explicit alignment supervision, with classification affected more than regression. \textit{w/o~$\mathcal{L}_{\text{recon}}$} shows the smallest degradation overall, mainly on EcoRegion ($-1.7$) and Housing ($-0.033$), suggesting that the reconstruction loss contributes most when fine-grained token specialization is required. Symmetric soft attention (\textit{sym.\ soft att.}) degrades across all tasks, indicating that peaked image-side attention provides a more effective alignment target than soft distributions on both paths.

\paragraph{Site and token count.}
\cref{fig:ablation} (right) shows sensitivity to site and token counts. Classification and regression both improve with site count and saturate near 4{,}096 sites. Doubling to 8{,}192 provides no further gain and slightly degrades regression. Token count peaks at $R = 64$ and declines for larger values. We hypothesize that too few tokens under-specify the semantic vocabulary, while too many dilute the alignment signal. We use 4{,}096 sites and 64 tokens in all subsequent experiments. We also ablate different site initialization strategies, and find that across our suite of geospatial tasks there is minimal difference (results in appendix).

\begin{figure*}[t!]
    \centering
    \includegraphics[width=0.8\linewidth]{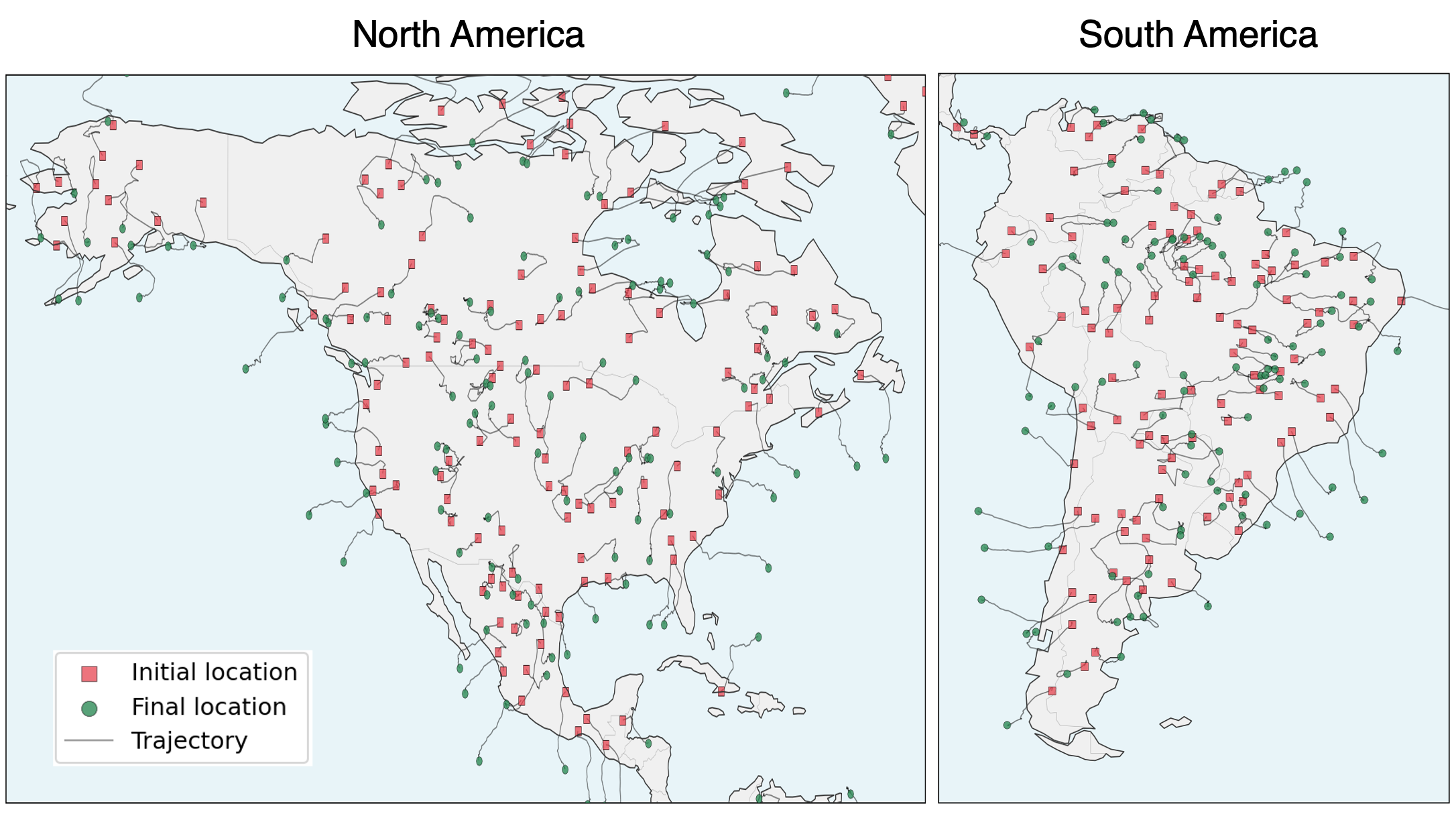}

    \caption{\textbf{Dynamics of Site Migration.} To maintain visual clarity, we visualize the training trajectories of a random 30\% subset of Voronoi sites. Markers indicate initial positions (\textcolor{Red}{$\blacksquare$}) and converged positions (\textcolor{ForestGreen}{$\bullet$}). Sites dynamically move toward geographic boundaries. Notably, many migrate offshore to specialize on narrow coastal segments, providing higher-resolution representational capacity where geographic features are densest.}
\label{fig:migration}
\end{figure*}

\subsection{Embedding Analysis}

\paragraph{Learned token semantics.}
\cref{fig:token_qualitative_quad} shows the token attention visualization for a few example tokens. Each panel displays the spatial attention map (which locations activate this token most strongly) alongside the satellite images with the highest attention scores. Individual tokens learn to specialize in semantically coherent environmental concepts. We observe tokens corresponding to snow and ice cover, arid desert, rocky terrain, and other visually distinct land types. The full set of token visualizations appears in the appendix.

\paragraph{Learned tessellation.} \cref{fig:tessellation} visualizes the learned spatial decomposition at three stages of the encoding pipeline. Panel (a) plots the positions of all 4{,}096 sites after training where we observe sites concentrating on land masses, clustering near coastlines and areas of discrimination while leaving oceans sparsely covered. Panel (b) assigns each land pixel to its nearest site and colors it by the ICA projection of that site's embedding vector, revealing that neighboring sites learn similar representations and form a coherent spatial code. Panel (c) shows the final location embeddings produced by the full model, including soft assignment and semantic token fusion, visualized via ICA. We see that embeddings vary smoothly across space. The non-uniform allocation in all three views emerges entirely from the contrastive training signal, without explicit supervision on site placement.

\paragraph{Site migration.}
\cref{fig:migration} visualizes site displacement over training for two regions. Sites are initialized on a Fibonacci lattice (red squares) and migrate during contrastive training to their converged positions (green circles), with trajectories shown as connecting lines. In North America, sites move from interior positions toward the coasts, the Appalachians, and the US-Mexico border region. In South America, sites converge along the boundaries of the Amazon Basin, the Andes, and the coastline. In both regions, sites in homogeneous interiors undergo little displacement, whereas those near geographic or ecological transitions migrate substantially, often moving offshore to specialize on narrow coastal segments rather than splitting capacity between the coast and the interior. We present additional site maps for different regions of the world in the appendix.
\section{Conclusion}
\label{sec:conclusion}

We introduced TTE, a location encoder that replaces fixed positional bases with learnable Spherical Voronoi partitions augmented by global semantic tokens. The Voronoi backbone provides an adaptive spatial decomposition with sites migrating during training to concentrate representational capacity in discriminative areas, while leaving homogeneous areas sparsely covered. Global semantic tokens complement this local spatial structure with a global semantic layer, distilling visual knowledge from the co-aligned satellite imagery into a compact vocabulary that enables geographically distant sites covering similar environments to share meaning through a common set of learned concept tokens. TTE sets a new state of the art among parametric location encoders, improving average classification accuracy by 2.8\% and average regression $R^2$ by 0.045 over the previous best on a suite of geospatial benchmarks, while achieving the strongest results as a geographic prior for fine-grained species classification on iNaturalist-2018. Beyond the quantitative results, the qualitative analysis reveals that TTE learns intuitive global semantic tokens specializing in coherent environmental concepts such as snow cover, arid desert, rocky terrain, and coastal areas. Voronoi sites discover a non-obvious geometric strategy, migrating offshore to dedicate their receptive fields to narrow coastal segments rather than splitting capacity between coast and interior. These behaviors emerge entirely from the contrastive training signal, suggesting that the combination of adaptive spatial partitioning and shared semantic anchors provides a rich inductive bias for geographic representation learning.

Several directions remain open. TTE currently trains on Sentinel-2 imagery without conditioning on time. Integrating multi-temporal observations would enable temporal knowledge accumulation, addressing limitations on temporally sensitive tasks such as housing price prediction. Incorporating additional modalities, such as ground-level photographs or environmental attributes could further improve performance on tasks that would benefit from them, such as fine-grained ecological classification. Finally, the Voronoi-plus-tokens architecture is not specific to geographic encoding. Any domain in which a function on the sphere exhibits non-uniform complexity, from planetary science to molecular surface modeling, could benefit from the same principle: learnable, adaptive partitioning combined with a shared semantic structure.

\section*{Acknowledgments}
This research used the TGI RAILs advanced compute and data resource, which is supported by the National Science Foundation (award OAC-2232860) and the Taylor Geospatial Institute.

%
%
\bibliographystyle{splncs04}
\bibliography{main}

\clearpage
\setcounter{page}{1}   

\title{Supplementary Material: Tessellating The Earth}
\titlerunning{Tessellating The Earth -- Supplementary}
\author{}              
\authorrunning{}       
\institute{}           
\maketitle

\appendix

\section{Implementation Details}
\label{sec:appendix_method}

\paragraph{Pretraining.}
We train on the S2-100K~\cite{klemmer2025satclip} preprocessed dataset of globally sampled Sentinel-2 multispectral images, using 13-band input at $256 \times 256$ resolution center-cropped to $224 \times 224$. We mimic SatCLIP~\cite{klemmer2025satclip} in our image pathway, except for the global semantic registers projection. Training of final models run for 300 epochs on 2$\times$ NVIDIA H100 GPUs using PyTorch DDP, completing in approximately 10 hours. The effective batch size is 6{,}384 (3{,}192 per GPU). All training is conducted in FP32.

The image encoder is a MoCo-pretrained ViT-Large/16~\cite{wang2022ssl4eo} with 309.9M parameters. TTE adds 4.0M trainable parameters (1.3\% of total) across 4{,}096 Voronoi sites (each with a 3D position, scalar temperature, and 384-dimensional embedding), 64 semantic tokens of dimension 512, and additional projection heads and MLP modules. Since both TTE and SatCLIP share the same frozen image encoder and differ only in the location encoder, TTE's total parameter count is less than 1\% larger than SatCLIP's, yet yields strong gains (Tab.~\mainref{1} in the main paper). 

\paragraph{Optimization.}
We use AdamW~\cite{loshchilov2017decoupled_adamw} with weight decay 0.01, per-parameter-group learning rates ranging from $5{\times}10^{-5}$ (projection heads) to $1{\times}10^{-3}$ (site positions), linear warmup for 5 epochs followed by cosine annealing to $10^{-6}$, and gradient clipping at L2 norm 1.0. We find lower temperatures on the site parameters leads to more optimal usage of tokens and smoother optimization overall.

\paragraph{Semantic token configuration.}
The 64 tokens use a register dropout rate of 0.1 during training. The location-path temperature is annealed linearly from 0.5 to 0.2 over training, while the image-path temperature is fixed at 0.05 to produce peaked attention. The fusion gate is fixed at $\alpha = 0.5$ (equal blend of Voronoi embedding and token-attended representation).

\paragraph{Evaluation protocol.}
We follow the RANGE~\cite{dhakal2025range} evaluation protocol. Embeddings are extracted from the frozen location encoder and min-max normalized to $[0, 1]$. For classification tasks (Biome, EcoRegion, Country), we use a ridge classifier with 10-fold cross-validation over $\alpha \in \{0.1, 1.0, 10.0\}$. For regression tasks (Temperature, Elevation, Population, California Housing), we use a ridge regression with 3-fold cross-validation over the same $\alpha$ set. All tasks use an 80/20 train/validation split with random seed 42. Final metrics are accuracy (classification) or $R^2$ (regression) on the held-out test set.

\paragraph{Site Parameter Details.}
Each site $k$ has the following learnable parameters:

\begin{table}[h]
\centering
\begin{tabular}{llll}
\toprule
Parameter & Symbol & Storage & Constraint \\
\midrule
Position & $\mathbf{s}_k \in \mathbb{S}^2$ & Unconstrained $\mathbb{R}^3$ & Normalized after each step \\
Temperature & $\tau_k > 0$ & $\log \tau_k \in \mathbb{R}$ & Clamped to $[\log 0.5, \log 500]$ \\
Embedding & $\mathbf{e}_k \in \mathbb{R}^D$ & Direct & None \\
\bottomrule
\end{tabular}
\end{table}

\section{Learned Parameter Visualizations}
\label{sec:appendix_visualizations}

\begin{figure*}
    \centering
    \begin{minipage}{0.6\linewidth}
        \centering
        \includegraphics[width=\linewidth,height=3.8cm,keepaspectratio]{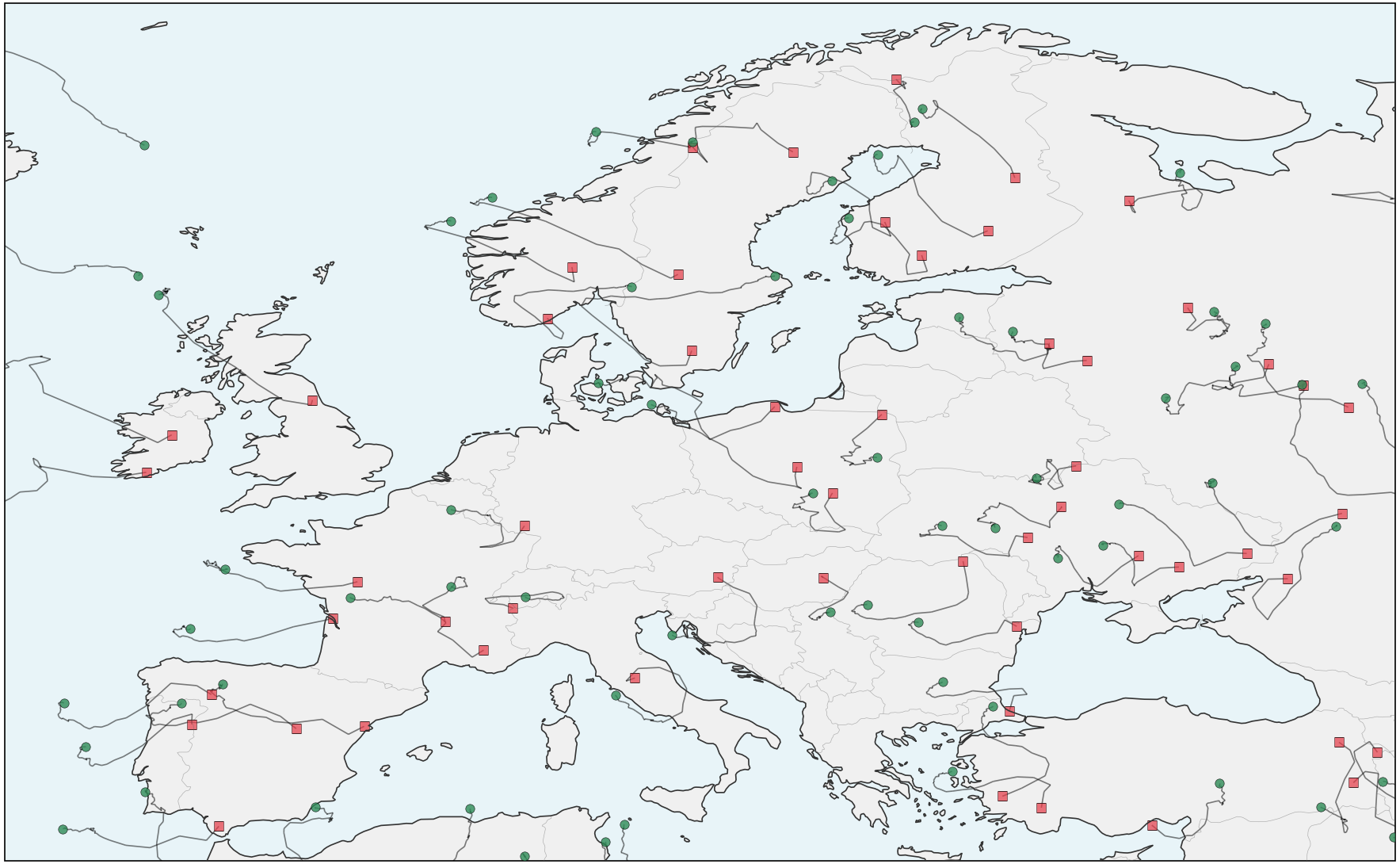}
    \end{minipage}%
    \hfill
    \begin{minipage}{0.36\linewidth}
        \centering
        \includegraphics[width=\linewidth,height=3.8cm,keepaspectratio]{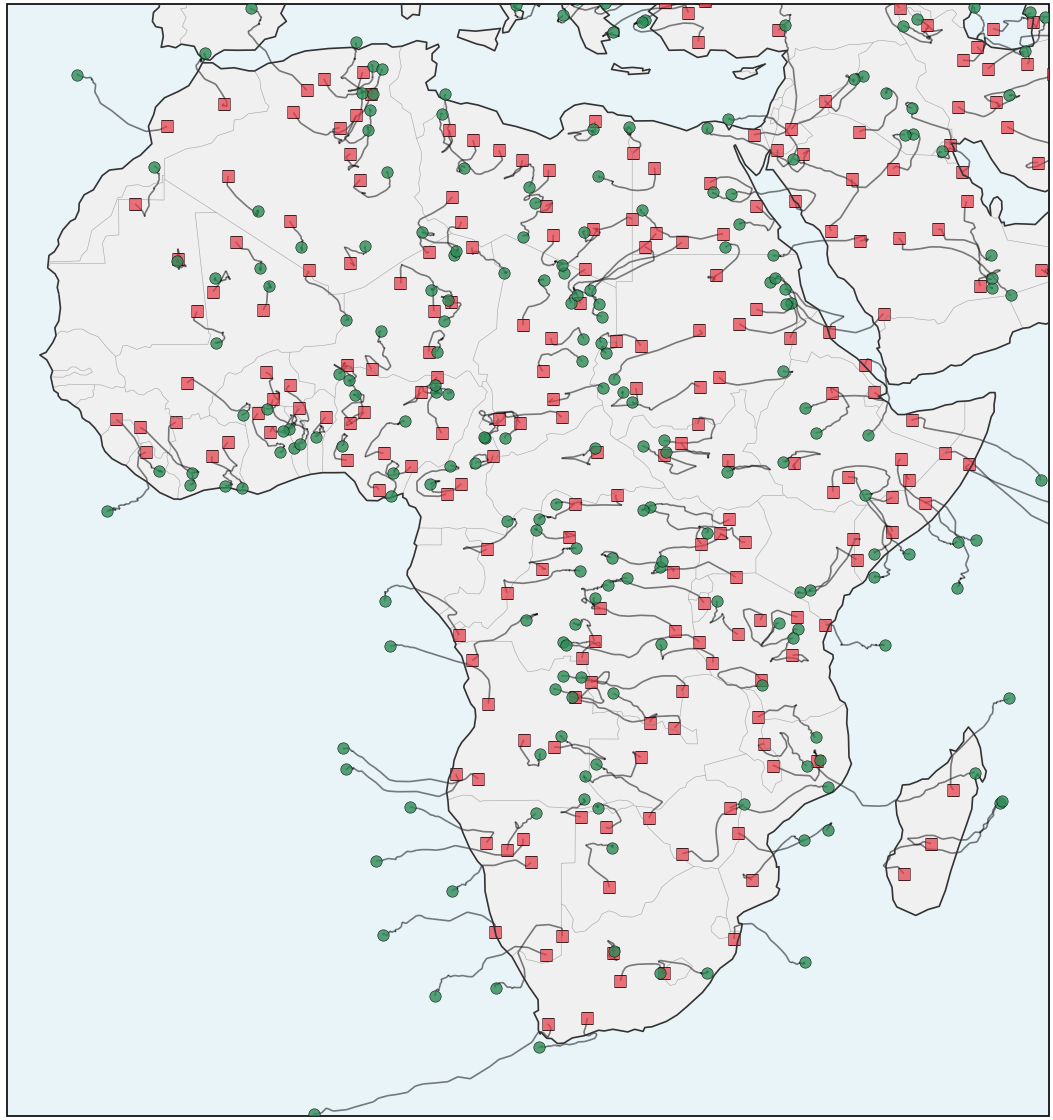}
    \end{minipage}

    \vspace{6pt}
    
    \begin{minipage}{0.6\linewidth}
        \centering
        \includegraphics[width=\linewidth,height=3.8cm,keepaspectratio]{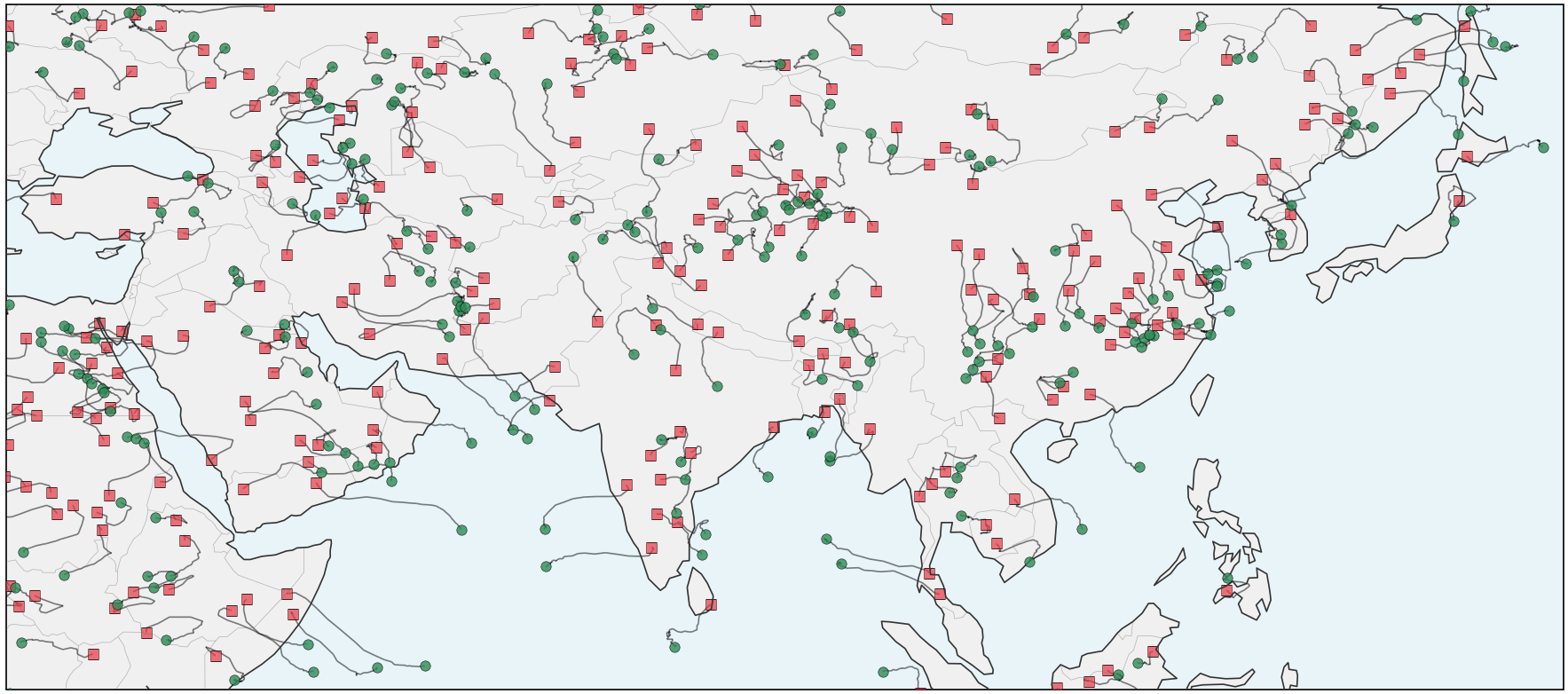}
    \end{minipage}%
    \hfill
    \begin{minipage}{0.36\linewidth}
        \centering
        \includegraphics[width=\linewidth,height=3.8cm,keepaspectratio]{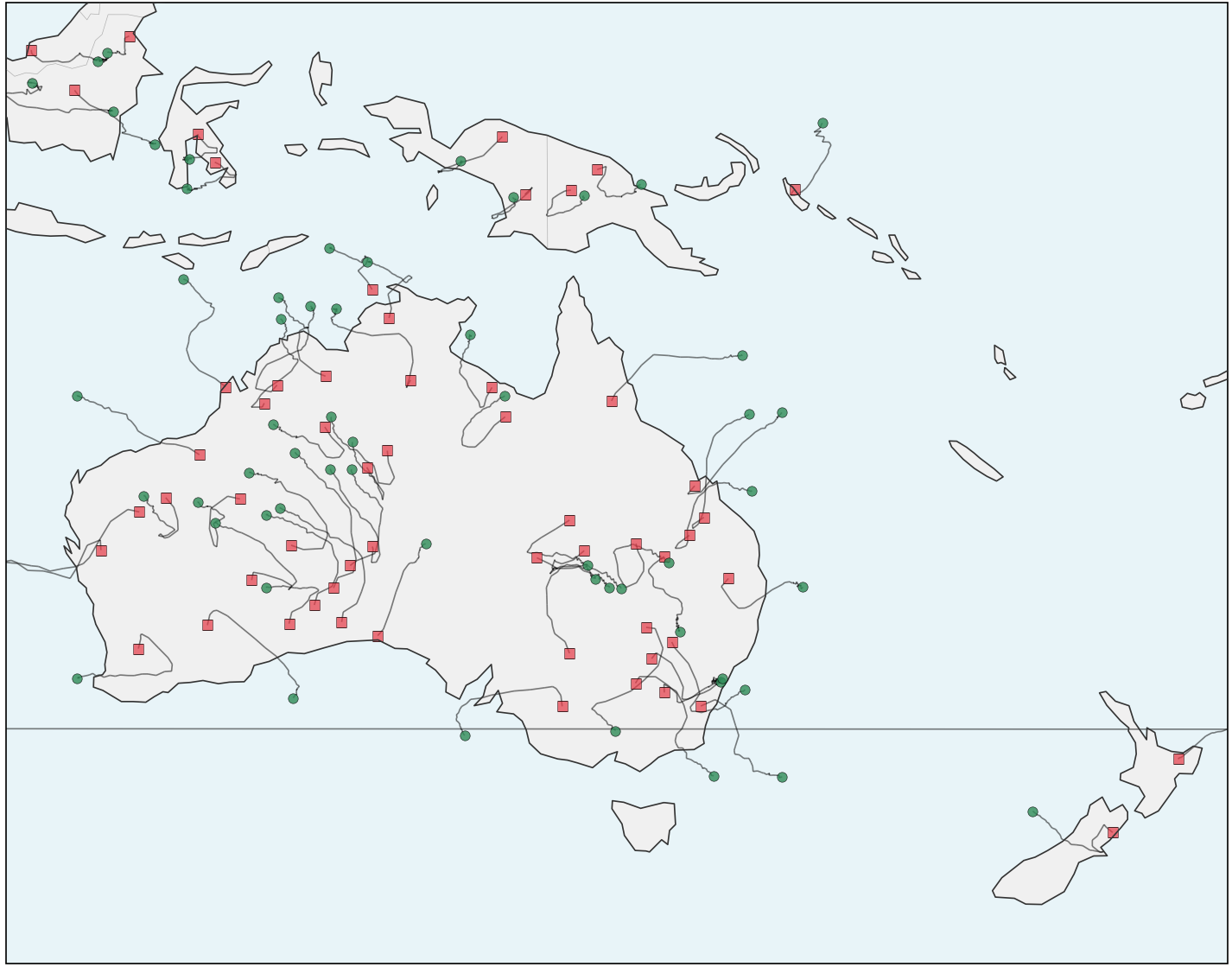}
    \end{minipage}
    \caption{\textbf{Site migration across regions.} Voronoi sites are initialized on a Fibonacci lattice. Markers indicate initial positions (\textcolor{Red}{$\blacksquare$}) and converged positions (\textcolor{ForestGreen}{$\bullet$}). Connecting lines show displacement. Random 30\% subset is shown.}
    \label{fig:migration_all}
\end{figure*}

\paragraph{All Region Site Migration.}
Fig.~\mainref{7} in the main paper visualizes site migration for North America and South America. \cref{fig:migration_all} extends this to additional regions. In each panel, markers indicate initial positions (\textcolor{Red}{$\blacksquare$}) and converged positions (\textcolor{ForestGreen}{$\bullet$}). Similar to the main paper, we see that sites move off the coast to focus on coastal boundaries, and otherwise move to discriminative areas.

\begin{figure}
    \centering
    \includegraphics[width=\linewidth]{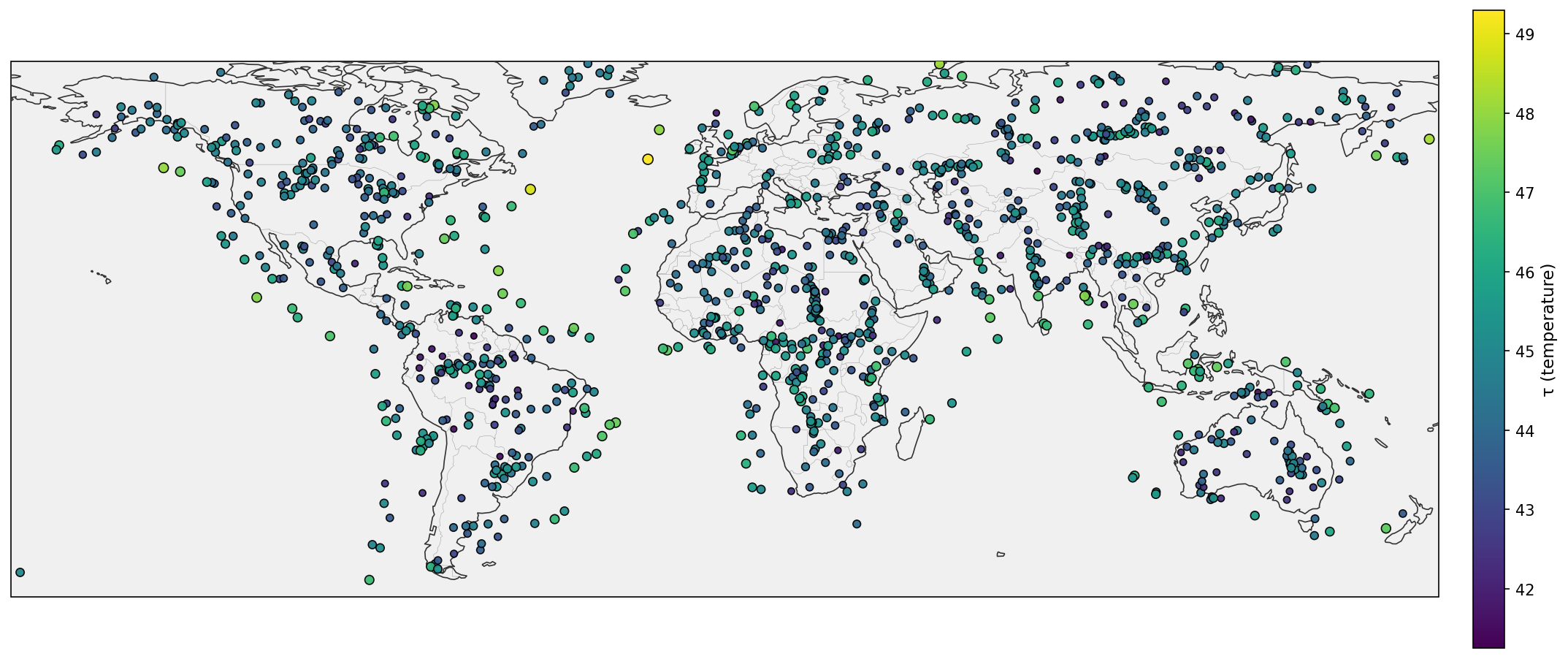}
    \caption{\textbf{Learned site temperatures.} Each site is plotted at its converged position and colored by its learned temperature $\tau_k$. Higher temperatures (warm colors) indicate sharper partition boundaries. We randomly select 30\% of sites for visual discrimination.}
    \label{fig:temperature_map}
\end{figure}

\paragraph{Learned Temperature Distribution.}
\cref{fig:temperature_map} visualizes the learned per-site temperature $\tau_k$ for a randomly selected 30\% of sites. Higher temperatures produce sharper Voronoi boundaries, effectively shrinking a site's receptive field and increasing local spatial resolution. Generally, we see that low-temperature sites are coupled with higher-temperature sites, which likely serve as more dominant embeddings of the general area. Isolated sites also have higher temperatures, likely indicating a more localized embedding.

\begin{figure*}
    \centering
    \includegraphics[width=0.6\linewidth]{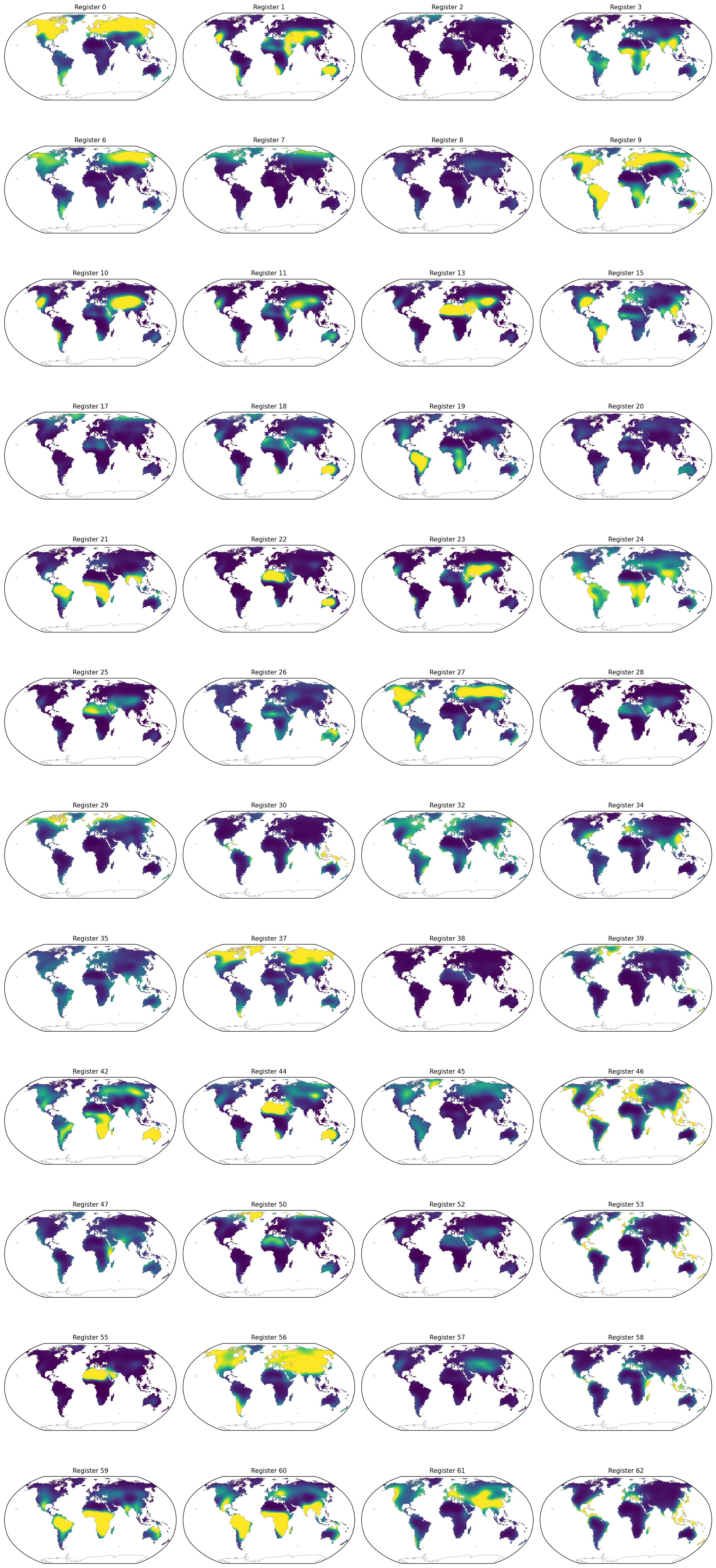}
    \caption{\textbf{Spatial attention maps of global semantic tokens.} Each panel shows one token's attention weight across the globe. Tokens specialize in diverse geographic and environmental patterns without explicit supervision. We select the  48 tokens with the most activations.}
    \label{fig:all_tokens}
\end{figure*}

\begin{figure}
    \centering
    \includegraphics[width=0.7\linewidth]{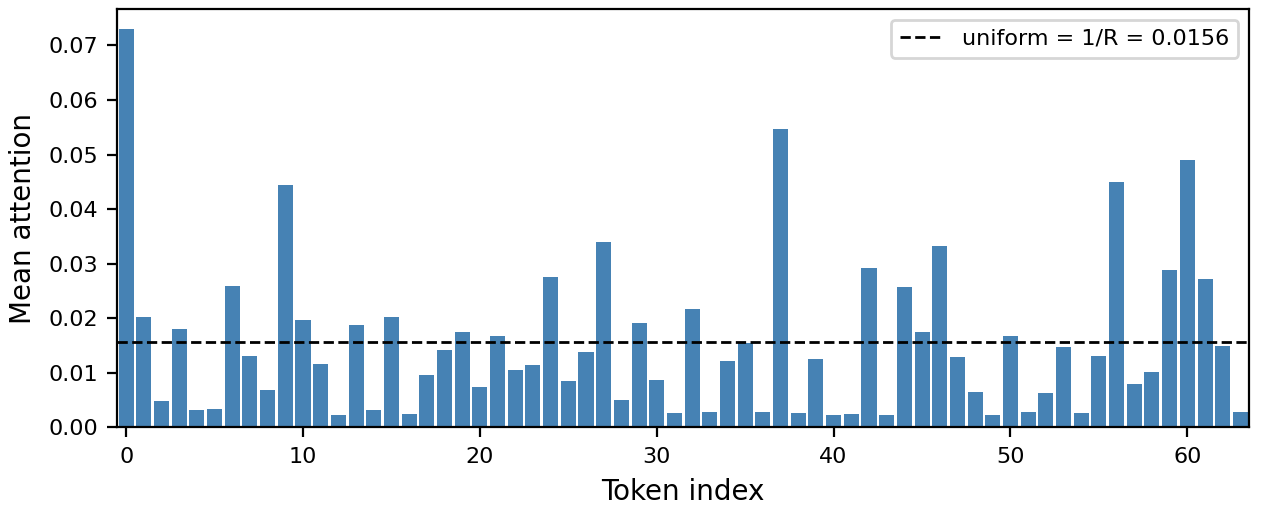}
    \caption{\textbf{Per-token mean attention.} Mean location-path attention received by each of the 64 semantic tokens, sorted in descending order, over a dense global query sample. $24/64$ tokens lie above the uniform baseline (dashed). The per-query attention entropy is $0.79$, corresponding to $\approx\!27$ effective tokens per query, indicating the encoder draws on a broad token vocabulary rather than collapsing onto a few.}
    \label{fig:per_token_usage}
\end{figure}

\paragraph{Global Semantic Tokens.}
Fig.~\mainref{3} in the main paper highlights four tokens with clear semantic specializations. \cref{fig:all_tokens} presents the spatial attention maps for 48 of the 64 tokens. Each cell shows the global attention weight of a single token across the sphere, revealing the geographic regions where that token is most active. We see many different specializations across the tokens indicating different semantic properties being captured.  The faint panels correspond to genuinely low-usage tokens that encode rarer environmental patterns. The encoder distributes attention across a broad vocabulary while still favoring the most frequently activated concepts. \cref{fig:per_token_usage} reports the mean location-path attention each token receives over a dense global query sample. Of the 64 tokens, 24 receive above-uniform mean attention, and the per-query attention entropy is 0.79, corresponding to roughly 27 effective tokens per query.

\section{Site Count}
\label{sec:appendix_data_scale}

\begin{figure}[h]
    \centering
    \includegraphics[width=\linewidth]{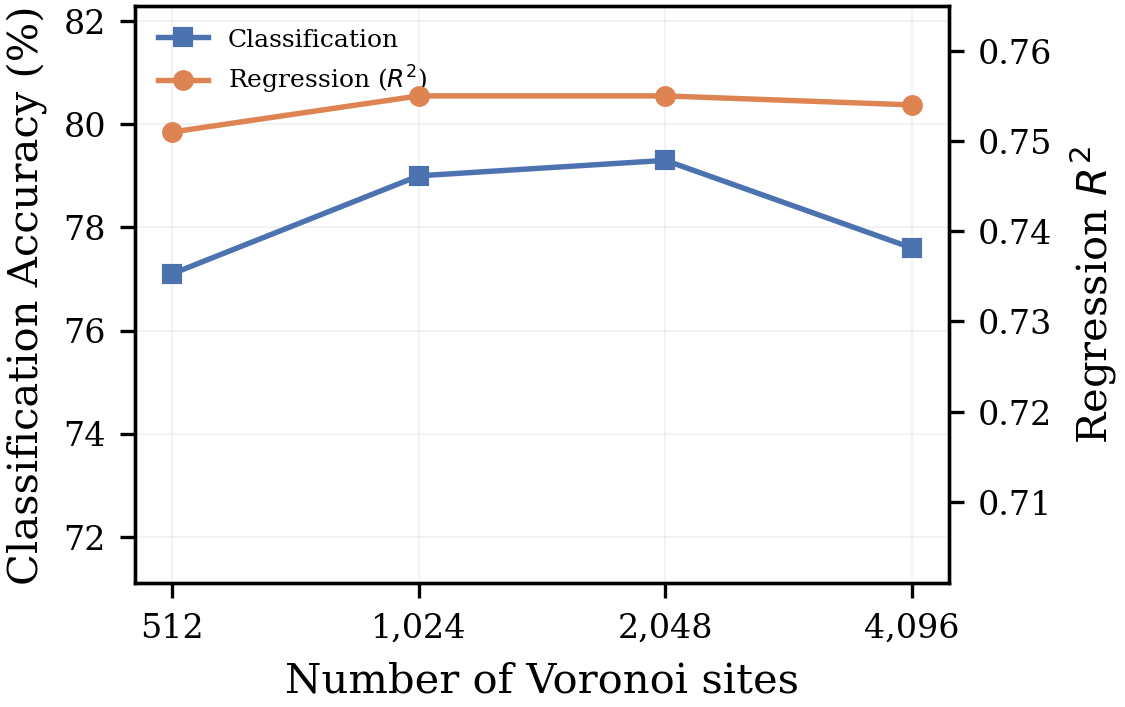}
    \caption{{Performance vs.\ Voronoi site count with reduced data.} All models trained at 50\% data scale for 70 epochs. Classification columns report accuracy (\%); regression columns report $R^2$.}
    \label{fig:half_data}
\end{figure}

\cref{fig:half_data} reports per-task performance across site counts at 50\% training data scale for 70 epochs. This supplements Fig.~\mainref{6} in the main paper by examining the effect of different data sizes on optimal site counts. 

Performance improves steadily up to $K{=}1{,}024$ and largely plateaus beyond that, with $K{=}2{,}048$ performing optimally on average. Increasing to $K{=}4{,}096$ provides no further gain and introduces slight degradations on some tasks, suggesting that additional site capacity at this data scale is not utilized effectively. This is consistent with the full-data results in the main paper, where $K{=}4{,}096$ is optimal. The reduced dataset simply shifts the capacity saturation point leftward.

\section{Effective Receptive Field}
\label{sec:appendix_topk}

\begin{figure}[h]
    \centering
    \includegraphics[width=\linewidth]{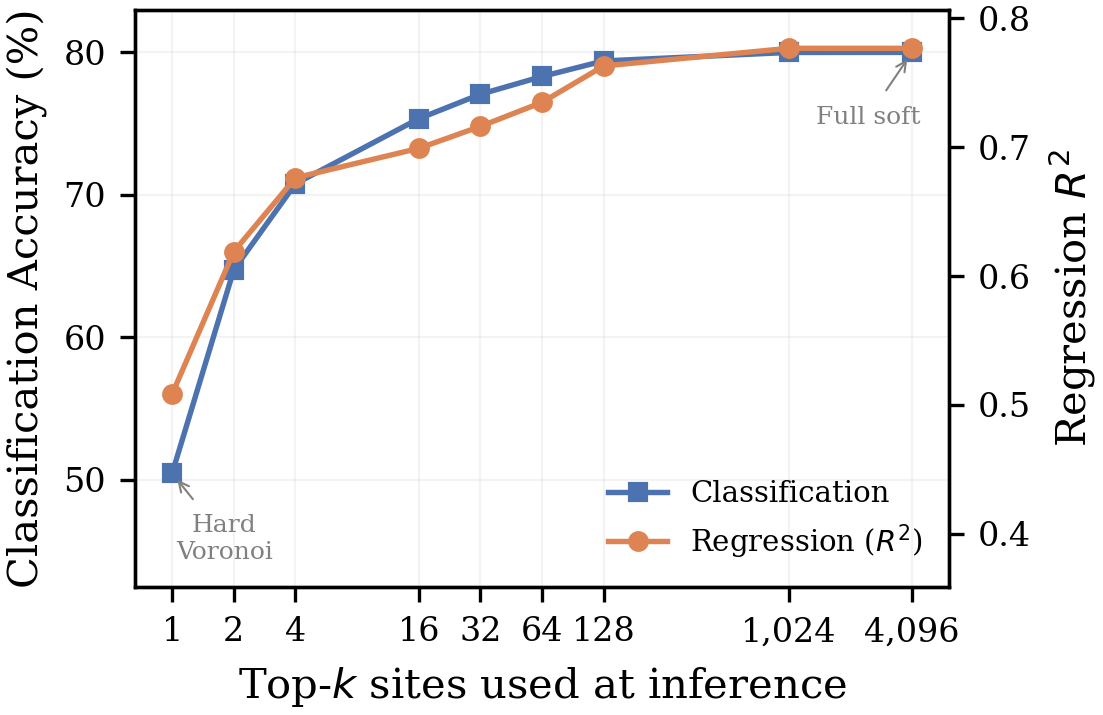}
    \caption{\textbf{Effective receptive field: top-$k$ site assignment at inference.} Average classification accuracy (left) and regression $R^2$ (right) as a function of the number of retained sites $k$. All evaluations use a single model trained with full soft assignment ($k{=}4{,}096$). Only the inference-time truncation varies.}
    \label{fig:topk_curve}
\end{figure}

TTE computes soft assignment weights over all $K = 4{,}096$ sites (Eq.~\mainref{2}, main paper), but in practice the per-site temperatures concentrate most weight on a small neighborhood of nearby sites. To characterize this effective receptive field, we evaluate a single trained model as we vary the number of sites used during inference. For a given budget $k$, we retain only the $k$ sites with the highest assignment weights for each query and renormalize their weights so that they sum to 1, zeroing out all others. At $k = 1$ this reduces to hard Voronoi assignment; at $k = K$ it recovers the full soft model. \cref{fig:topk_curve} shows performance as a function of $k$. We notice that K=128 shows minimal degradation.

\section{Additional Ablations}
\label{sec:appendix_additional}

\paragraph{Site Initialization.}
The default TTE model initializes sites on a Fibonacci lattice filtered to land areas. We test whether initializing sites at positions derived from existing location encoders improves convergence or final performance (\cref{tab:additional_ablations}). Specifically, we cluster the embedding spaces of SatCLIP~\cite{klemmer2025satclip} and RANGE~\cite{dhakal2025range} into $K = 4{,}096$ clusters using $k$-means on a dense global sample, and use the cluster centroids (projected onto $\mathbb{S}^2$) as initial site positions.

\begin{figure*}
    \centering
    \begin{minipage}{0.32\linewidth}
        \centering
        \includegraphics[width=\linewidth]{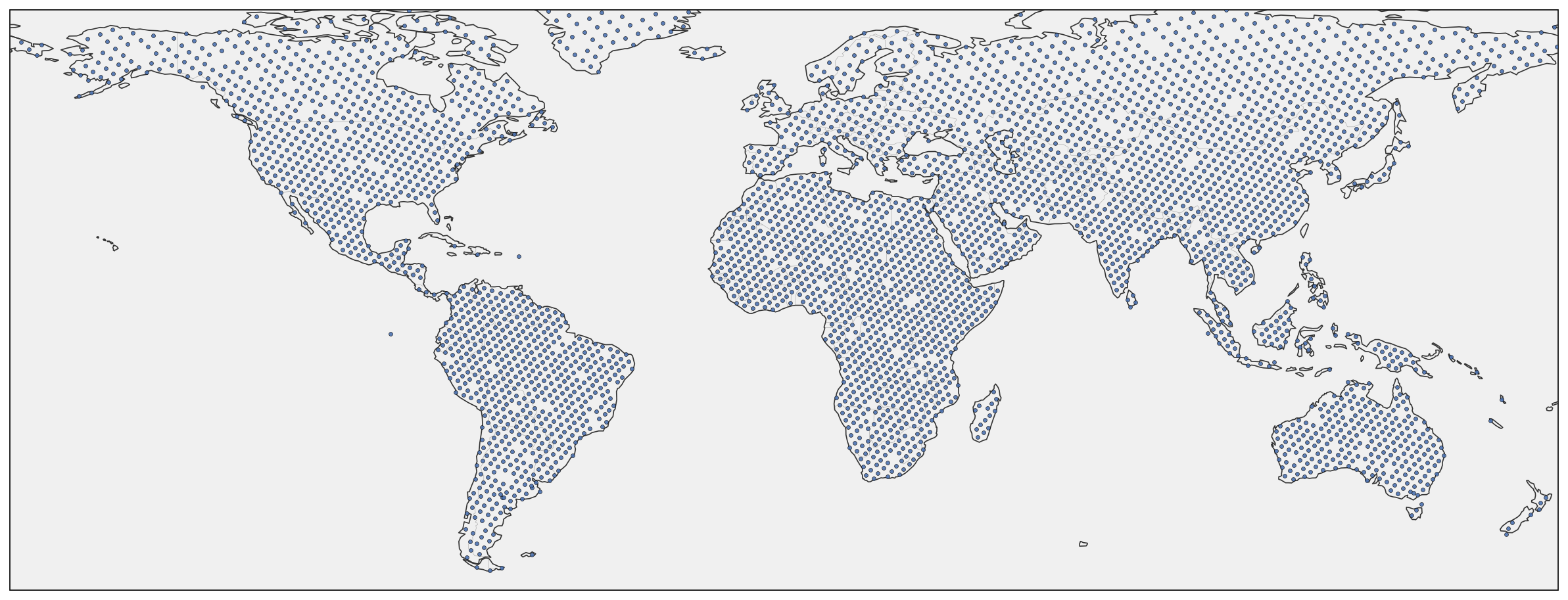}
        {\scriptsize (a) Fibonacci lattice}
    \end{minipage}%
    \hfill
    \begin{minipage}{0.32\linewidth}
        \centering
        \includegraphics[width=\linewidth]{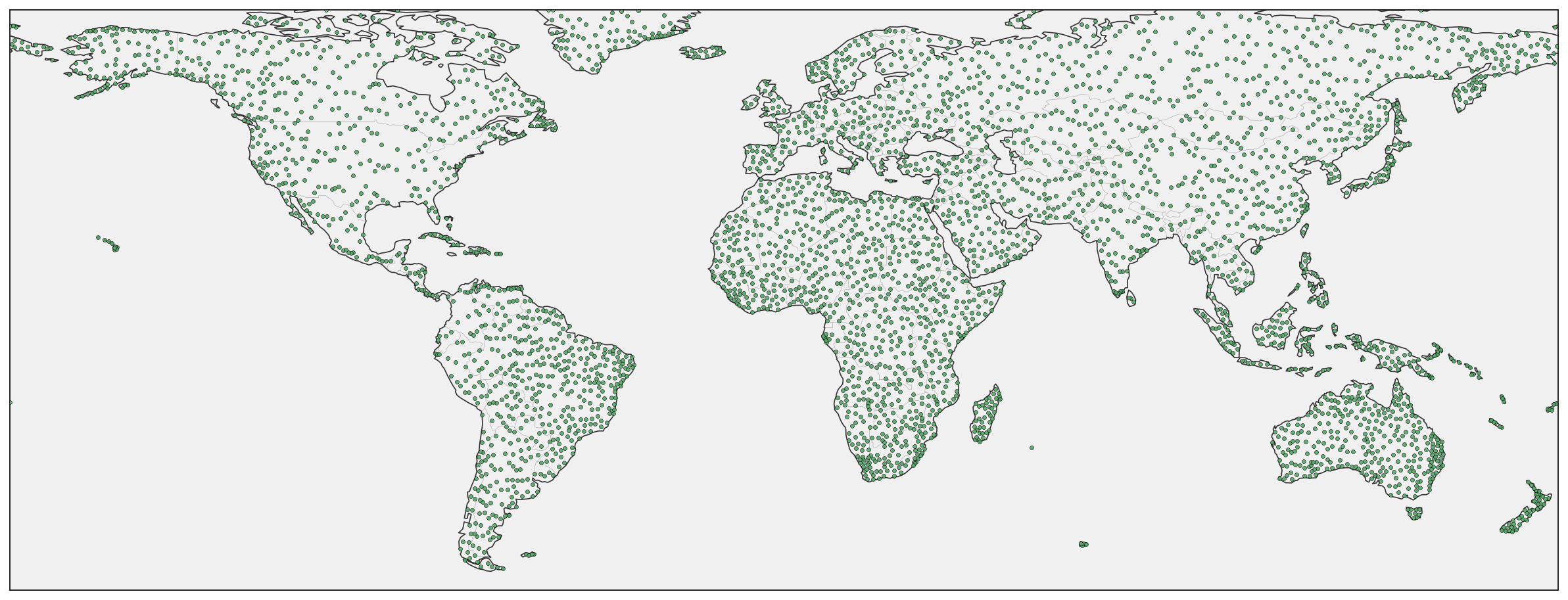}
        {\scriptsize (b) SatCLIP clusters}
    \end{minipage}%
    \hfill
    \begin{minipage}{0.32\linewidth}
        \centering
        \includegraphics[width=\linewidth]{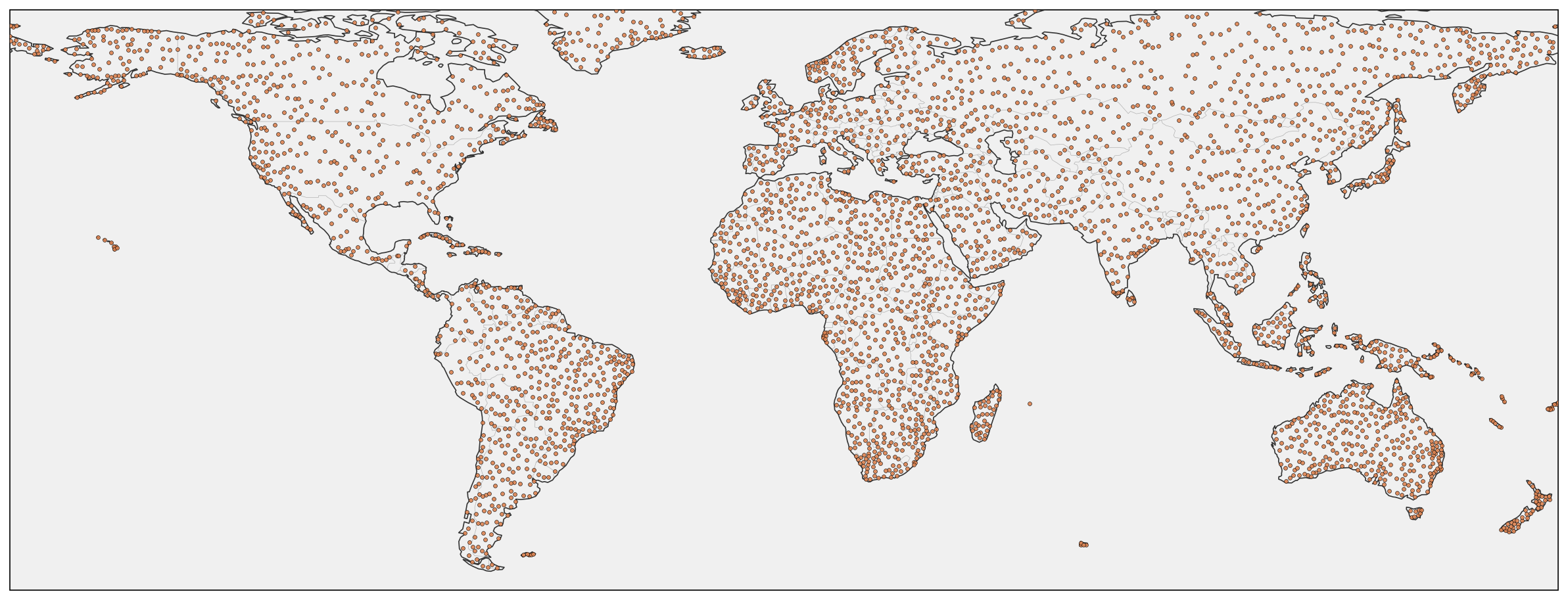}
        {\scriptsize (c) RANGE clusters}
    \end{minipage}
    \caption{\textbf{Site initialization strategies.} Initial site positions for each initialization method. (a)~Fibonacci lattice filtered to land areas distributes sites approximately uniformly. (b,c)~Cluster centroids derived from SatCLIP and RANGE embeddings concentrate sites in regions of high embedding complexity.}
    \label{fig:init_maps}
\end{figure*}

\paragraph{Coordinate Conditioning and 3D Sites.}

We test two architectural variants (\cref{tab:additional_ablations}). \textit{Coordinate conditioning} encodes the query coordinate with sin-cos encoding and concatenates the resulting features to the Voronoi embedding before passing through the residual MLP. This early fusion gives the MLP direct access to the coordinate conditioning alongside the discrete Voronoi representation. \textit{3D sites} removes the re-normalization step that projects site positions back to $\mathbb{S}^2$ after each gradient update, allowing sites to move off the unit sphere and gain a radial degree of freedom.

Coordinate conditioning produces mixed results. It improves EcoRegion (+2.5) and California Housing (+0.112) but degrades Biome ($-6.1$), Country ($-3.8$), and Elevation ($-0.172$). The gains on EcoRegion and Cali-Housing likely reflect the higher-frequency features providing signal that help the linear probe on tasks with sharp boundaries or localized spatial structure, but the regressions on other tasks suggest that the additional input interferes with the Voronoi encoder's learned partitioning. A hybrid approach that selectively applies coordinate conditioning more directly may warrant future investigation. 3D sites perform slightly below the constrained model across the board (Cls 79.0 vs.\ 80.0; Reg 0.765 vs.\ 0.777), confirming that the radial degree of freedom provides no useful capacity when training data is confined to a fixed sphere. 

\begin{table*}
\centering
\scriptsize
\caption{\textbf{Additional ablations.} \textit{Top:} site initialization strategies. \textit{Bottom:} architectural variants. All models use $K{=}4{,}096$ sites.}
\setlength{\tabcolsep}{3pt}
\begin{tabular}{l|ccc|c|cccc|c|cc}
\hline
& Bio & Eco & Ctry & Cls $\uparrow$
& Temp & Elev & Pop & Cali & Reg $\uparrow$
& iNat-1 & iNat-3 \\
\hline
TTE
& 77.8 & 67.4 & 94.8 & 80.0
& 0.946 & 0.839 & 0.790 & 0.532 & 0.777
& 76.2 & 90.0 \\
\hline
\multicolumn{12}{l}{\textit{Site initialization}} \\
\quad SatCLIP clusters
& 77.8 & 67.9 & 94.6 & 80.1
& 0.944 & 0.829 & 0.784 & 0.535 & 0.773
& 76.2 & 90.0 \\
\quad RANGE clusters
& 77.8 & 67.8 & 94.4 &80.0
& 0.948 & 0.831 & 0.785 & 0.512 & 0.769
& 76.2 & 90.0 \\
\hline
\multicolumn{12}{l}{\textit{Architectural variants}} \\
\quad TTE + coord.\ cond.
& 71.7 & 69.9 & 91.0 & 77.2
& 0.891 & 0.667 & 0.724 & 0.644 & 0.732
& 75.6 & 89.8 \\
\quad TTE + 3D sites
& 76.5 & 66.5 & 94.1 & 79.0
& 0.947 & 0.832 & 0.783 & 0.498 & 0.765
& 75.9 & 89.8 \\
\hline
\end{tabular}
\label{tab:additional_ablations}
\end{table*}

\section{Inference-Time Augmentation}
\label{sec:appendix_inference}

\begin{table*}
\centering
\scriptsize
\caption{\textbf{Inference-time retrieval augmentation on geospatial tasks.} TTE as a standalone parametric encoder (top) compared with RANGE applied to TTE (middle) and RANGE applied to SatCLIP as reported by Dhakal~\etal~\cite{dhakal2025range} (bottom). Classification columns report accuracy; regression columns report $R^2$.}
\setlength{\tabcolsep}{4pt}
\begin{tabular}{l|ccc|c|cccc|c}
\hline
& \multicolumn{3}{c|}{\textbf{Classification}}
&
& \multicolumn{4}{c|}{\textbf{Regression}}
& \\
\cline{2-4} \cline{6-9}
\textbf{Configuration}
& Bio & Eco & Ctry & Cls $\uparrow$
& Temp & Elev & Pop & Cali & Reg $\uparrow$ \\
\hline
TTE (parametric only)
& 77.8 & 67.4 & 94.8 & 80.0
& 0.946 & 0.839 & 0.790 & 0.532 & 0.777 \\
\hline
\multicolumn{10}{l}{\textit{RANGE applied to TTE}} \\
\quad semantic
    & \textbf{84.2} & 73.8 & \textbf{97.5} & \textbf{85.2}
    & 0.958 & \textbf{0.871} & \textbf{0.819} & \textbf{0.555} & \textbf{0.801} \\
\quad semantic + geodesic
    & 83.4 & 73.1 & 95.8 & 84.1
    & \textbf{0.964} & \textbf{0.871} & \textbf{0.819} & 0.551 & \textbf{0.801} \\
\hline
SatCLIP (parametric only)
& 68.9 & 69.3 & 82.8 & 73.7 
& 0.825 & 0.666 & 0.684 & 0.400 & 0.644 \\
\hline
\multicolumn{10}{l}{\textit{RANGE applied to SatCLIP~\cite{klemmer2025satclip,dhakal2025range}}} \\
\quad RANGE
    & 83.3 & \textbf{75.7} & 93.7 & 84.2
    & 0.895 & 0.844 & 0.799 & 0.422 & 0.740 \\
\quad RANGE+
    & 83.3 & 75.3 & 94.7 & 84.4
    & 0.931 & 0.851 & 0.811 & 0.336 & 0.732 \\
\end{tabular}
\label{tab:range_comparison}
\end{table*}

RANGE~\cite{dhakal2025range} augments a parametric location encoder with image features retrieved from an external database at inference time. Because RANGE operates under fundamentally different deployment assumptions than purely parametric methods, requiring storage of and access to an external image feature database, we report this comparison separately from the main results.

RANGE retrieves visual features using two complementary signals: \emph{semantic} similarity (cosine similarity between the query location embedding and database location embeddings) and \emph{geodesic} proximity (angular distance between the query and database coordinates on the sphere). The geodesic component enforces spatial smoothness by upweighting features from nearby locations regardless of their embedding similarity. We apply RANGE on top of TTE using both retrieval modes and compare against the published RANGE results built on SatCLIP~\cite{klemmer2025satclip}.

\begin{table}
\centering
\caption{\textbf{iNaturalist-2018 comparison with retrieval-augmented methods.} Top-$k$ classification accuracy (\%) on the iNaturalist-2018 test split. TTE as a purely parametric encoder outperforms RANGE and RANGE+, which require an external image database at inference time, across all $k$.}
\setlength{\tabcolsep}{4pt}
\begin{tabular}{l|cccc}
\hline
 & Top-1 & Top-3 & Top-5 & Top-10 \\
\hline
Img                & 66.1 & 83.3 & 88.0 & 92.2 \\
\hline
Img + CSP~\cite{mai2023csp}          & 72.9 & 87.9 & 91.6 & 94.8 \\
Img + GeoCLIP~\cite{cepeda2023geoclip}      & 72.9 & 88.2 & 91.9 & 95.2 \\
Img + CSP INat$^\ast$~\cite{mai2023csp}  & 74.4 & 88.8 & 92.2 & 94.9 \\
Img + SatCLIP~\cite{klemmer2025satclip}        & 75.1 & 88.7 & 91.9 & 94.5 \\
Img + TaxaBind~\cite{sastry2025taxabind}       & 75.1 & 89.7 & 93.0 & 95.6 \\
\hline
Img + RANGE~\cite{dhakal2025range}        & 75.2 & 89.6 & 92.9 & 95.5 \\
Img + RANGE+~\cite{dhakal2025range}       & 75.1 & 89.5 & 92.8 & 95.5 \\
\hline
\rowcolor{gray!15}
Img + TTE & \textbf{76.2} & \textbf{90.0} & \textbf{93.2} & \textbf{95.7} \\
\hline
\end{tabular}
\label{tab:inat_range}
\end{table}

\paragraph{Geospatial tasks.}
\cref{tab:range_comparison} presents the geospatial results. Several patterns emerge.

Semantic RANGE applied to TTE achieves the highest classification average (85.2) and regression average (0.801) of any configuration, outperforming RANGE applied to SatCLIP on Country (+3.8), Temperature (+0.063 over RANGE), Elevation (+0.027), and California Housing (+0.133). RANGE on SatCLIP retains an advantage on EcoRegion (75.7 vs.\ 73.8), likely reflecting SatCLIP's denser fixed-grid coverage in ecologically complex regions.

Adding geodesic proximity to the retrieval signal yields benefits over SatCLIP on some tasks (RANGE+ vs.\ RANGE), most notably for Temperature ($0.931$ vs.\ $0.895$) and Country ($94.7$ vs.\ $93.7$). On top of TTE, however, geodesic retrieval slightly degrades classification (Biome: $84.2 \to 83.4$, Country: $97.5 \to 95.8$) while leaving regression unchanged. This is consistent with TTE's learned site structure. We hypothesize that, because sites migrate during training to areas of discrimination identified by the satellite imagery, the encoder already captures the local spatial structure that geodesic retrieval provides for fixed-basis encoders. Thus, geodesic neighbors of a TTE embedding are likely to carry redundant information, and their inclusion can introduce noise.

Purely semantic retrieval provides substantial gains over standalone TTE across all tasks (e.g., Biome: $77.8 \to 84.2$, EcoRegion: $67.4 \to 73.8$, Elevation: $0.839 \to 0.871$), indicating that the 64 semantic tokens do not fully capture the richness of direct image features. The remaining gap represents visual information available in satellite imagery that cannot be compressed into a fixed token vocabulary using our approach.

\paragraph{iNaturalist-2018.}
\cref{tab:inat_range} extends the comparison to fine-grained species classification. It is notable that TTE as a purely parametric encoder outperforms both RANGE ($75.2$) and RANGE+ ($75.1$) across all Top-$k$ thresholds, despite requiring no external database at inference time.

\end{document}